\documentclass[letterpaper, 10 pt, conference]{ieeeconf}
\pdfoutput=1
\usepackage[dvipsnames,HTML]{xcolor}
\usepackage{subcaption}
\DeclareCaptionSubType*[arabic]{figure}
\usepackage{graphicx}
\usepackage{epsfig} % for postscript graphics files
\usepackage{amsmath, amssymb}
\usepackage{textcomp} % Pro­vide many text sym­bols baht, bul­let, copy­right, ...
\usepackage{gensymb} % Pro­vides generic com­mands \de­gree, \cel­sius, \pert­hou­sand, \mi­cro and \ohm which work both in text and maths mode.
\usepackage{multirow}
\usepackage{array}
\usepackage{flushend}
\usepackage{sidecap}
\usepackage{color}
\usepackage[normalem]{ulem}

\usepackage[utf8]{inputenc}
\DeclareUnicodeCharacter{0252}{\"u}
\DeclareUnicodeCharacter{0246}{\"o}
\DeclareUnicodeCharacter{0237}{\'\i}
\DeclareUnicodeCharacter{0225}{\'a}

\usepackage{verbatim}
\usepackage{arydshln}
\usepackage[parfill]{parskip}
\usepackage[export]{adjustbox}
\usepackage[colorlinks=true, citecolor=blue]{hyperref}

\title{\LARGE \bf Human Visual Understanding for Cognition and Manipulation - A primer for the roboticist}

\author{Martin Hjelm}

\begin{document}

\maketitle

\begin{abstract}
Robotic research is often built on approaches that are motivated by insights from self-examination of how we interface with the world. However, given current theories about human cognition and sensory processing, it is reasonable to assume that the internal workings of the brain are separate from how we interface with the world and ourselves. To amend some of these misconceptions arising from self-examination this article reviews human visual understanding for cognition and action, specifically manipulation. 

Our focus is on identifying overarching principles such as the separation into visual processing for action and cognition, hierarchical processing of visual input, and the contextual and anticipatory nature of visual processing for action. We also provide a rudimentary exposition of previous theories about visual understanding that shows how self-examination can lead down the wrong path. 

Our hope is that the article will provide insights for the robotic researcher that can help them navigate the path of self-examination, give them an overview of current theories about human visual processing, as well as provide a source for further relevant reading.
\end{abstract}

\begin{comment}
1. state the problem
2. say why it is interesting
3. say what your solution achieves
4. say what follows from your solution.
\end{comment}

% \input{todo}
\section{Introduction}

A major source of inspiration for works in robotics has been the
examination of different organisms, including humans. In modeling
robotic cognition for human environments self-examination has been a
guiding component. Self-examination can, however, be a double-edged
sword.

\begin{figure}[ht!]
% l b r t
\centering
\includegraphics[width=.75\linewidth,clip,trim=355 250 325 225]{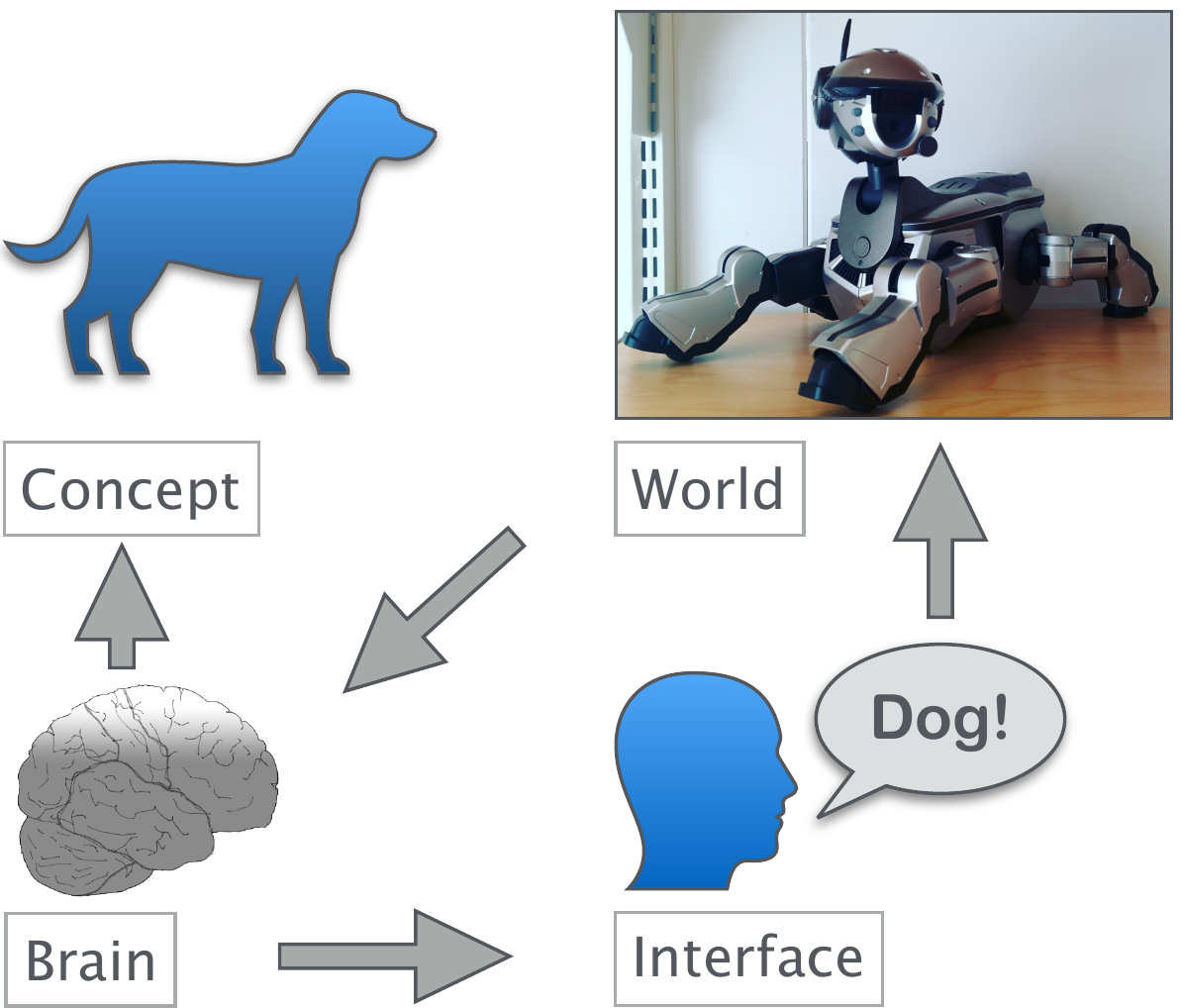}
\caption{\label{fig:brain_interface_model}Brain-World Interaction: One way of approaching the brain-world interactions is by thinking of language as an interface for the world. The brain receives sensory input, interprets it dependent on the context and the number of sensory inputs, mapping it into a concept. The concept is then mapped into some relevant form in our consciousness be it a word or awareness or an action command. The brain does not, for example, count features of the object it sees and then decides what it is seeing.}
\end{figure}

In robotics, vision has been the major modality used for cognition and
decisions regarding action. As vision is one of our primary senses it is
easy to draw conclusions from self-examination of how we, ourselves,
cognize about what we see. Human vision research is full of examples of
theories that use ideas grounded in the workings of language and how we
interface with the world. However, current models of human vision show
to the contrary that the processing of visual input is significantly
different from how we use vision and language to interface with the
world. Fig. \ref{fig:brain_interface_model} gives a simple illustration
how one can approach the intersection of language, vision, the world.

It is impossible to account for something as complex as human visual
understanding within one article. We will, therefore, give a rough
outline of current ideas of the human brain's object image processing
system tied together with models of internal object representation and
how that connects to action, specifically to grasping. As in-depth
expositions already exist \cite{Chinellato:2015uj, Kruger:2013wg} our
aim is to give a readily available source explaining main principles,
strategies, and the structure of human visual understanding without too
much of for the novice, confusing nomenclature used in psychological and
neuroscience research.

\begin{figure}[ht!]
% l b r t
\centering
\includegraphics[width=.99\linewidth,clip,trim=180 380 335 245]{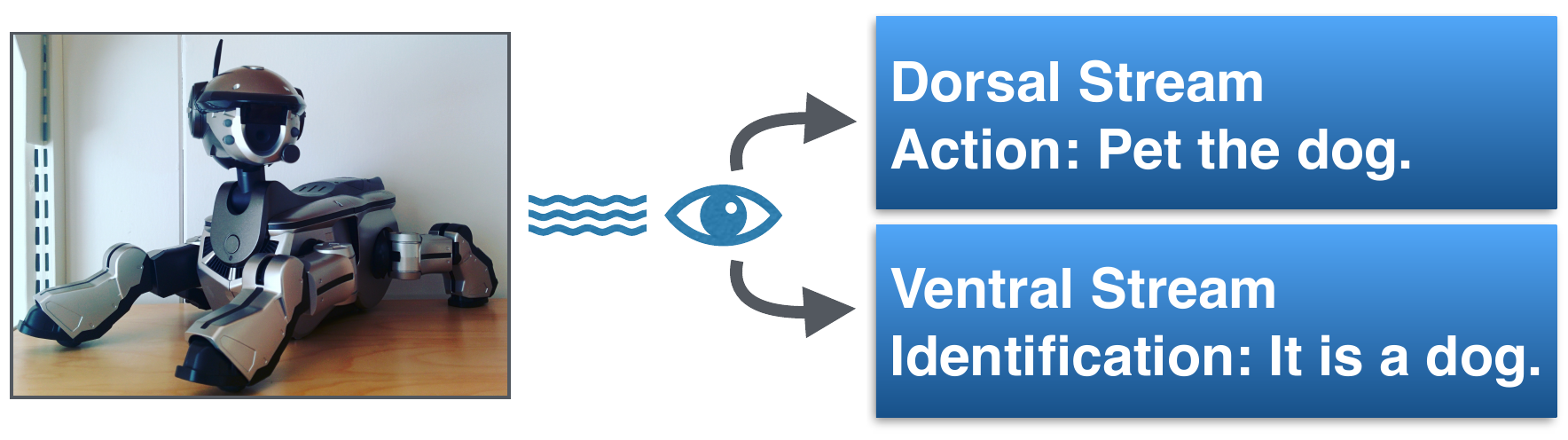}
\caption{\label{fig:dorsal_ventral_sketch} Simplified sketch of the division of labor between the dorsal and ventral visual processing stream. The ventral pathway is generally considered to answer questions about \textit{what} we see, while the dorsal pathway answers questions about \textit{how} we can perform an action on what we see.}
\end{figure}

We start with a review of human visual processing following the standard
model of the two processing pathways one for semantic object recognition
and one for action, fig.\ref{fig:dorsal_ventral_sketch}. We then proceed
to give a short review of prior models of human visual understanding
comparing to what self-examination tells us. We compare these models to
the current models of humans vision, that is, recurrent, hierarchical
processing of visual input into an increasingly invariant and linearly
separable feature space, and show how it accounts for the previous
models. Finally, we review vision for action. We show how the object
meaning influences action representation and how that representation
affects the grasping process. We also review how intention, context, and
the post-grasp task influences human grasping and the visual processing.

Worth noting is that much of the understanding of the human brain comes
from research done on monkeys, and it is often assumed that there are
homologs, that is, equivalent functional parts, in the human brain. We
will follow this idea and try to be clear about from where the
experimental evidence comes as this is not always evident in the
literature.

\begin{figure}
% l b r t
\centering
\includegraphics[width=1.\linewidth,clip,trim=280 225 180 190]{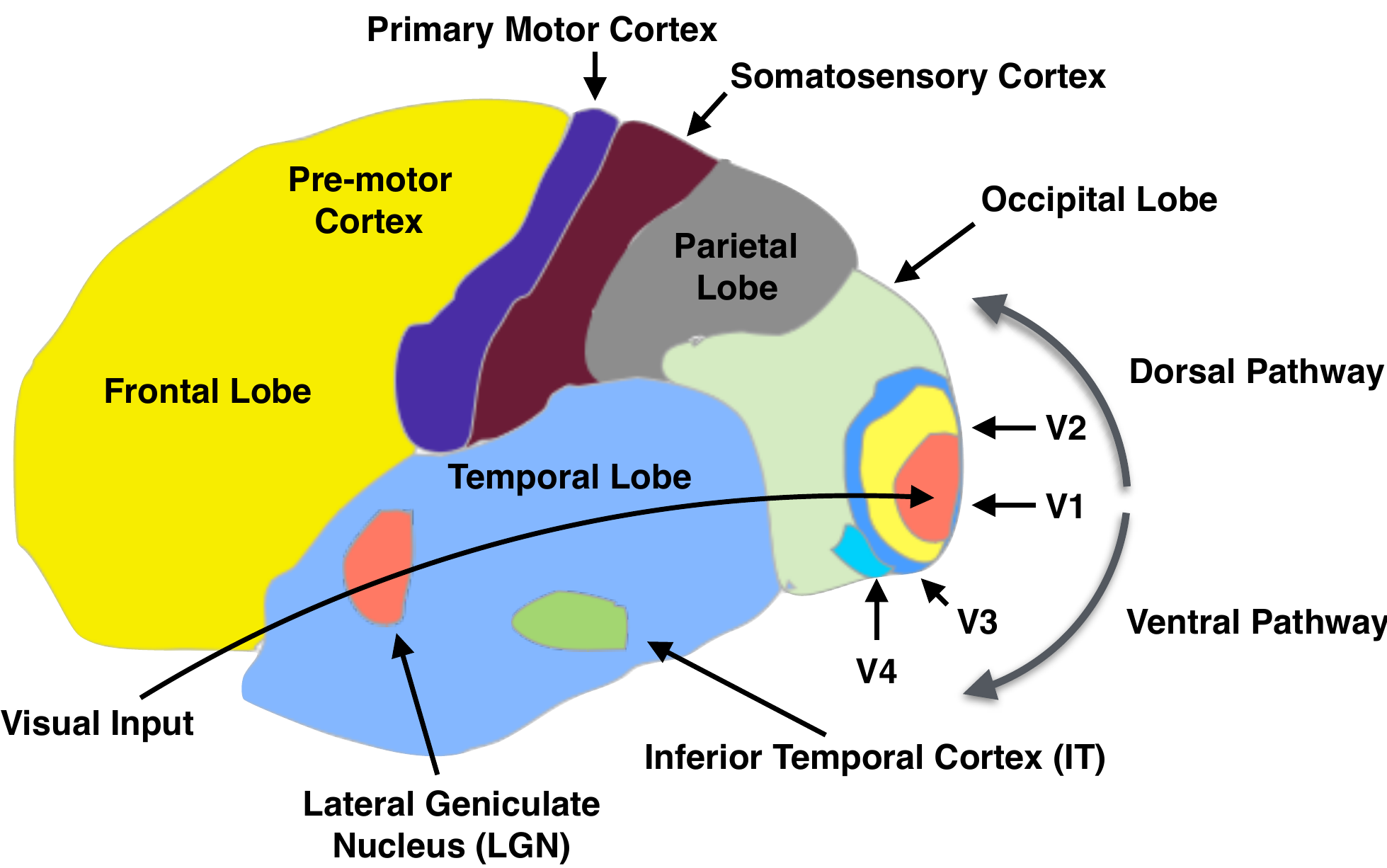}
\caption{\label{fig:dorsal_ventral_brain} Rough sketch of the visual processing in the brain based on afigure in \cite{Logothetis:1999ub}. The visual input passes through the LGN which combines the two visual streams and forwards it to the back of the brain, the Occipital lobe, where the primary visual cortex is located. The visual information flows through area V1 and V2 and is then split into the dorsal and ventral pathways. The dorsal pathway is thought to process visual information related to action and is sent to the parietal lobe for integration with other sensory information and onwards to the primary motor cortex for execution of actions. The ventral pathway ends in the Inferior Temporal Cortex and is believed to process the visual information in a hierarchical and increasingly complex fashion to facilitate semantic understanding of the visual input.}
\end{figure}

\section{Human Visual Processing}

The most influential model of human image processing is that of two
anatomically and functionally distinct pathways, the dorsal and ventral
pathways \cite{Goodale:1992gq} (fig.\ref{fig:dorsal_ventral_sketch}).
The dorsal named the ``how'' pathway links to the motor system and is
thought to encode spatial information needed for interaction with the
world. The ventral stream, named the ``what'' pathway, is thought to
encode information for object recognition and general visual perception.

The reason for the \emph{``how''} label of the dorsal pathway is that
lesions, damaged tissue in the brain, in the ventral and dorsal cortex
of primates cause degradation in object perception or spatial vision
\cite{Ungerleider:1982uz}. Additional evidence for the distinction
between the two pathways was a human subject, D.F., with lesions in
parts of the ventral pathway. D.F. had impaired object recognition and
could not recognize line-drawings of objects but showed normal
pre-shaping and rotation of the hand when grasping objects implying that
location, orientation, size, and shape estimation of objects was intact
\cite{James2463, Goodale:1992gq}. \cite{Goodale:1992gq} put forward that
both streams process the attributes of objects that are manipulated, but
for different purposes.

The distinction between the two pathways should be viewed as a
simplified model of visual processing in the brain. There is plenty of
evidence of information sharing between the two. Much of the information
encoded in each pathway is used by the other in such things as shape
perception, object detection, and intentional visuomotor action
\cite{Farivar:2009tt, Grafton:2010jo, Lebedev:2002kh, Milner:2003io, Schenk:2010jq, Schenk:2011uh, Zachariou:2014cl}.

We can outline a general model of the current understanding of visual
processing under the two pathways as follows
(fig.\ref{fig:dorsal_ventral_sketch}). The visual input from the retinas
is sent to the lateral geniculate nucleus (LGN) which combines the input
from the two retinas and their different receptors. It forwards the
visual stream to the primary visual cortex (V1) which processes features
such as orientation, direction, and color. V1 forwards the information
to area V2 where it splits into the two pathways the ventral and dorsal.

The ventral pathway proceeds into area V4 and the inferior-temporal (IT)
cortex which is considered the end of the ventral pathway. The IT
contains several areas that activate during visual input of among others
faces, body parts, scenes, and different shapes.

The dorsal pathway continues onto area V3 and the caudal intraparietal
area (CIP), and then onto the anterior intraparietal cortex (AIP),
ending in the primary motor cortex (M1), thus transforming visual input
into action. Apart from the feed-forwarding the visual processing
pathways also contain feedback projections allowing earlier visual areas
to receive processed information from the later stages
\cite{Kravitz:2013uk}.

\begin{figure*}[!ht]
    \centering %left,bottom,right,top
    \includegraphics[width=1.\textwidth,clip,trim=120 180 180 100]{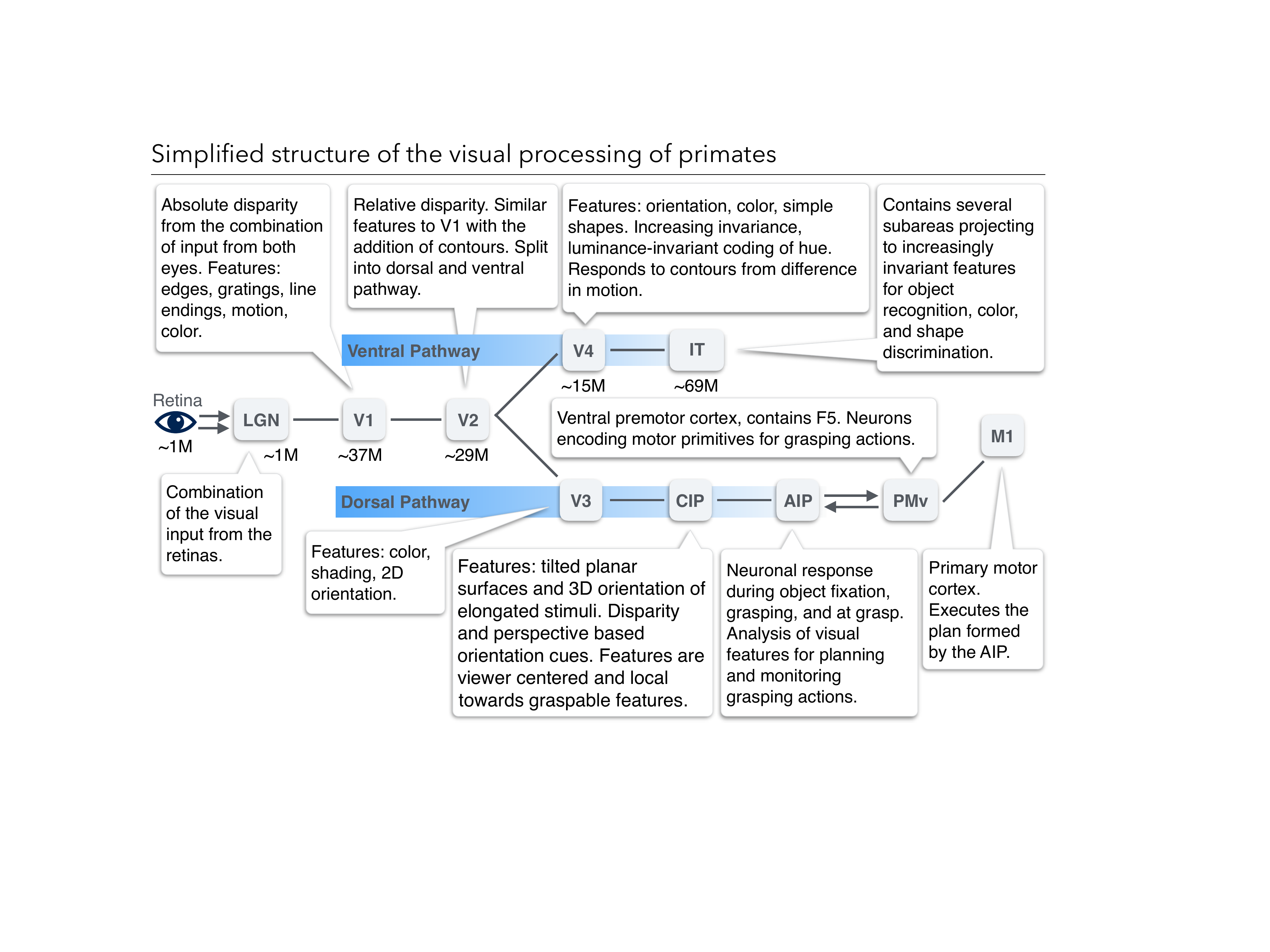}
    \caption{Simplified structure of the visual processing of primates aggregated from \cite{Chinellato:2015uj, Kruger:2013wg, Grafton:2010jo, Castiello:2005gq, Logothetis:1996tr, DiCarlo:2012em}. The model is extremely simplified and the visual processing connects to many more areas than given here, in addition, there are plenty of recurrent connections between the areas. For in-depth models see the referred papers. An approximate number of neurons per area is shown beneath each area box for the ventral pathway (from \cite{DiCarlo:2012em}). Visual input flows from the two retinas to the LGN where it is combined into one visual stream. The primary visual cortex (V1) computes features such as edges, gratings, line endings, motion, color, and absolute disparity. The features in V2 are similar to the ones in V1 with the addition of simpler contours and computation of relative disparity. In V2 the visual stream is split into the dorsal and ventral pathways. The ventral stream goes onto V4 which computes features such as orientation, color, and simple shapes. There is an increasing invariance and a luminance-invariant coding of hue. V4 also responds to contours from difference in motion. V4 projects to the inferior temporal cortex (IT) which is responsible for among other things object categorization. The dorsal stream continues on to V3 which computes color, shading, and 2D orientation. This information is forwarded to the CIP which computes surface orientations, and disparity based and perspective based orientation cues. Features are viewer-centered, local, and directed towards graspable features. The information in the CIP is forwarded to the AIP which together with the PMv, which contains movement primitives for grasping movements, plans the grasping action.}
    \label{fig:brain_image_proceesing}
\end{figure*}

We illustrate and explain the flow of visual input and processing more
thoroughly in fig.\ref{fig:brain_image_proceesing}.

\subsection{The Ventral Pathway}

The ventral stream, the \emph{``what''} pathway, is considered to
encode, among other things, information related to object identity, that
is, both category and specific object identity. The word stream
indicates a feed-forward visual processing network. The predominant idea
has been that the network creates a hierarchy of more abstract
representations further down the stream which are increasingly invariant
to translation and rotation. However, there is plenty of evidence of
connections between the early and later stages of visual processing that
indicates that visual processing is a process of continuous refinement.

Recent models, therefore, point towards a recurrent and highly
interactive network. A review \cite{Kravitz:2013uk} suggest that the
network links information from at least six cortical and subcortical
systems such as early visual areas and areas responsible for high-level
representations. The feedforward and feedback projections allow for
efficient communication between adjacent areas but also the early and
late stages of the visual processing. Evidence of sensitivity to
retinotopic position even for higher level representations indicates
that visual recognition is an ongoing process of contextual calibration,
that is, from directing attention, to categorical and individual
recognition, to putting the recognition into its proper context. These
recurrent connections are also suggested as an explanation for the
brain's redundancy, that is, the retained functionality even after
extensive damage to certain areas \cite{Kravitz:2013uk}.

\subsection{Human Object Representations}

Models of human object representation range from neuropsychological
explanations that try to tie neurological evidence with cognition to
visual neuroscience that focuses on explaining vision through neural
activity. Different models have different strengths in explaining the
various phenomena in vision. Neuropsychological models make
simplifications of the processing in the ventral-dorsal stream to
explain cognitive abilities. Visual neuroscience models, on the other
hand, typically focus on explaining one specific phenomenon. Accounting
for all models and phenomena as well as experimental evidence is beyond
this paper. We will, therefore, limit the focus to models concerning
central abilities of human object recognition - the ability to
categorize, abstract, and identify - starting with some of the major
models from neuropsychology.

A full model of human vision needs to explain viewpoint invariance at
the basic category level, that is, invariance under the transformation
of retinal position, scale, luminance, deformation, clutter, context,
etc. It also needs to account for reduced viewpoint invariance for novel
objects. Experiments on humans and monkeys \cite{Logothetis:1996tr} show
that recognition drops for view disparities of depth-rotated objects
larger than \(30\degree\). However, when presented with two views
\(75\degree-120\degree\) apart the monkeys interpolated them to give
almost perfect recognition for any view between the two.

Further on, a full model also needs to explain categorization,
abstraction of objects into invariant features, and the generalization
into prototypical representations of objects in a category. Finally, it
needs to account for different levels of categorization for one and the
same object. This means, for example, being able to explain our ability
to discriminate between, canine, dog, Terrier, and a Terrier named
Rocky.

The abstraction of object features means decomposing objects into
meaningful entities, categories of their own, and their spatial
relations, an internal ontology. These subordinate categories are not
necessarily necessary for categorization of the object itself --- but as
discussed above --- are important for reasoning and conveying
information about a category.

Identification means the ability to match visual input to a specific
object from memory. To complicate matters object identity can be viewed
as a sub-category with cardinality one
\cite{Riesenhuber:2000bw, DiCarlo:2012em}. If we would have one shared
representation for all object recognition it would need to account for
all the category memberships from superordinate to identity.

\begin{figure*}[ht!]
    % r b l t
    \centering
    \includegraphics[width=.99\linewidth, clip,trim=98 410 240 160]{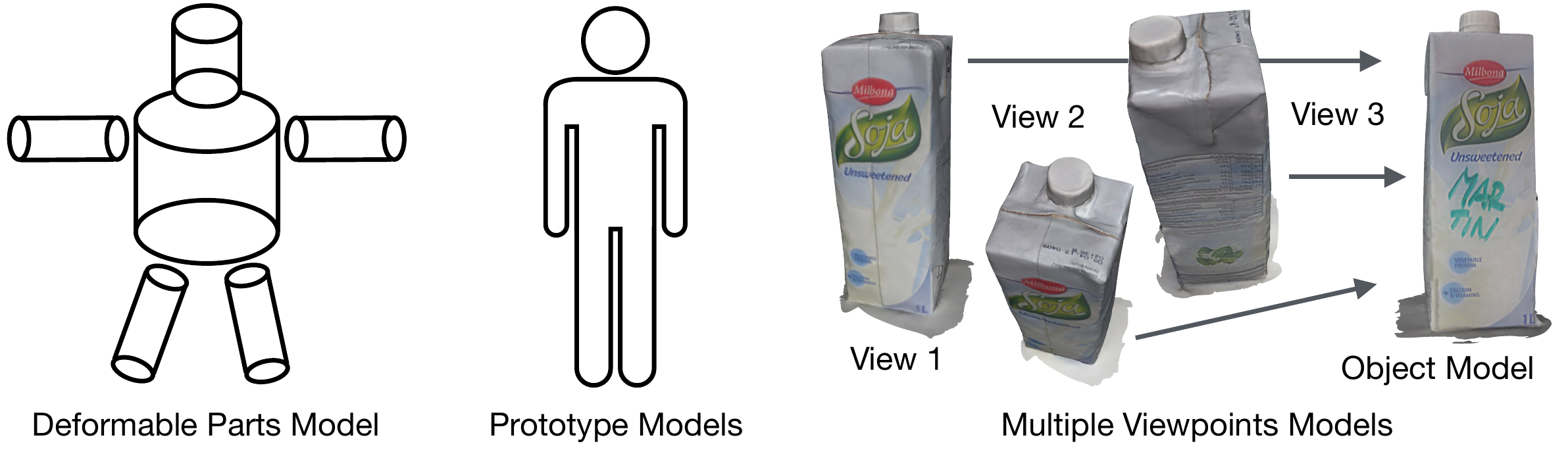}
    \caption{Models of visual understanding. \textbf{Deformable parts models}, models objects into smaller invariant parts where the relational structure is preserved, but where the underlying metric between the parts is not. Here a human is represented as a set of cylinders in a relational structure. \textbf{Prototype models}, models objects as the summarization of within-category objects into a generalized model. These prototypes can take one different granularity depending on the need e.g. pictogram vs. a more human-like model. \textbf{Mulitple viewpoints models}, models objects as the association or stitching together of multiple viewpoints of an object. A novel object is then matched to a viewpoint cluster.}
    \label{fig:visual_understanding_models}
\end{figure*}

\subsubsection{The Invariant Decomposable Parts Perspective}

\label{subsubsec:decomposable_perspective} Early theories of vision
modeled objects as decomposable into smaller invariant parts that
preserved the relational structure, but not the underlying metric
between the parts \cite{Tarr:2002bn}. Biederman \cite{Biederman:1987bx},
for example, suggested decomposing objects into geons, deformable,
prototypical parts such as cones, cuboids, cylinders, etc., that
together with a relational structure can represent objects
(fig.~\ref{fig:visual_understanding_models}).

The decomposable part models as initially formulated were not tied to
any experimental data \cite{Palmeri:2004bi, Peissig:2007dt}. We should
see them in the context of how we as humans reasons and abstract objects
to communicate about them. The major criticism of the decomposable part
model is its inability to account for rotation, that is, invariant
features and the relational structure might not be detectable from a
view occluding them but the object might still be identifiable. A
decomposable part based model must also account for the level of
granularity in the decomposition it needs to categorize the object
correctly which introduces additional complexity
\cite{Logothetis:1996tr}. If we simplify and consider the deformable
parts model a tree graph then the number of possible graphs for \(n\)
nodes will grow exponentially as \(n^{n-2}\). Finally, further
complications arise from the segmentation and relational structuring
having to happen interchangeably as the parts need to be recognized and
fit into a relational structure.

Ideas of decomposition are not uncommon in computer vision and have been
applied with varying success. Common approaches are Bag-of-Words models
\cite{Csurka04visualcategorization} and deformable parts models
\cite{Felzenszwalb:2010ez}. Works in grasping have also used
decomposition strategies. \cite{Miller:2003ev} represented objects as
decomposable into a set of geometric primitives where each primitive has
known pre-planned grasps. And \cite{Aleotti:2011hc} represents objects
as graphs where they associate each node with a set of pre-learned
grasps.

\subsubsection{The Multiple Viewpoints Perspective}

A more recent model of human object recognition, supported by
psychophysical and physiological data, is that of multiple viewpoints
\cite{Tarr:2002bn, Riesenhuber:2000bw}. It opposes the ideas of
invariance and structure arguing that the brain instead store
representations of objects as a set of unique viewpoints
(fig.~\ref{fig:visual_understanding_models}).

The major criticism of the multiple viewpoints model is that a small
perturbation may cause a new viewpoint to significantly differ from
previously stored viewpoints. Therefore, to form complete
representations that can account for recognition of novel viewpoints and
mental rotations, the viewpoints must be normalized and stitched
together in some fashion \cite{Tarr:2002bn}. Another shortcoming is
memory capacity. Unless there is some generalization, filtering, or
compression process it is unclear which viewpoints should be stored for
future use. The recognition process is also difficult as the brain,
during recognition, needs to match a viewpoint to viewpoint clusters of
objects retained in memory.

\subsubsection{The Prototype Perspective}

A third related model is that of prototypes
(fig.~\ref{fig:visual_understanding_models}), that is, the summarization
of within-category objects into a generalized model
\cite{Edelman:1995vd}. We can trace this idea as far back as to Plato.
To recognize an object, the visual input is matched to prototypes that
are kept in memory, using an invariant distance measure. The prototype
model of vision has some support in experimental data. For example,
experiments involving distortions of a set of simple patterns show that,
when infants and adults are given sufficient exemplars of a category,
they tend to abstract these into prototypical patterns
\cite{Logothetis:1996tr}.

Prototype models for vision are good at explaining abstractions. Yet,
similarly to the decomposable parts models, they are suspiciously close
to how we abstract and communicate about categories. In fact,
experiments on categorization in monkeys and humans show that the
strategies involved in subordinate categorization tasks are most likely
explained by referring to category exemplars rather than prototype
similarity \cite{Palmeri:2004bi}.

Prototypes models have problems in accounting for scaling, rotation, and
translation when matching the image to stored prototypes as these
operations are all very taxing for biological systems
\cite{Rolls:2012dt}. Prototypes are also inherently coded for
generalization. This means that a prototype based system will use many
different prototypes for different generalizations under different
contexts ranging from the ability to depict individual exemplars to
summarizing basic categories. This leads to the same problems faced by
the multiple viewpoints model an explosion in the number of prototypes.

\begin{figure}[ht!]
% r b l t
\centering
\includegraphics[width=1.\linewidth,clip,trim=380 390 280 135]{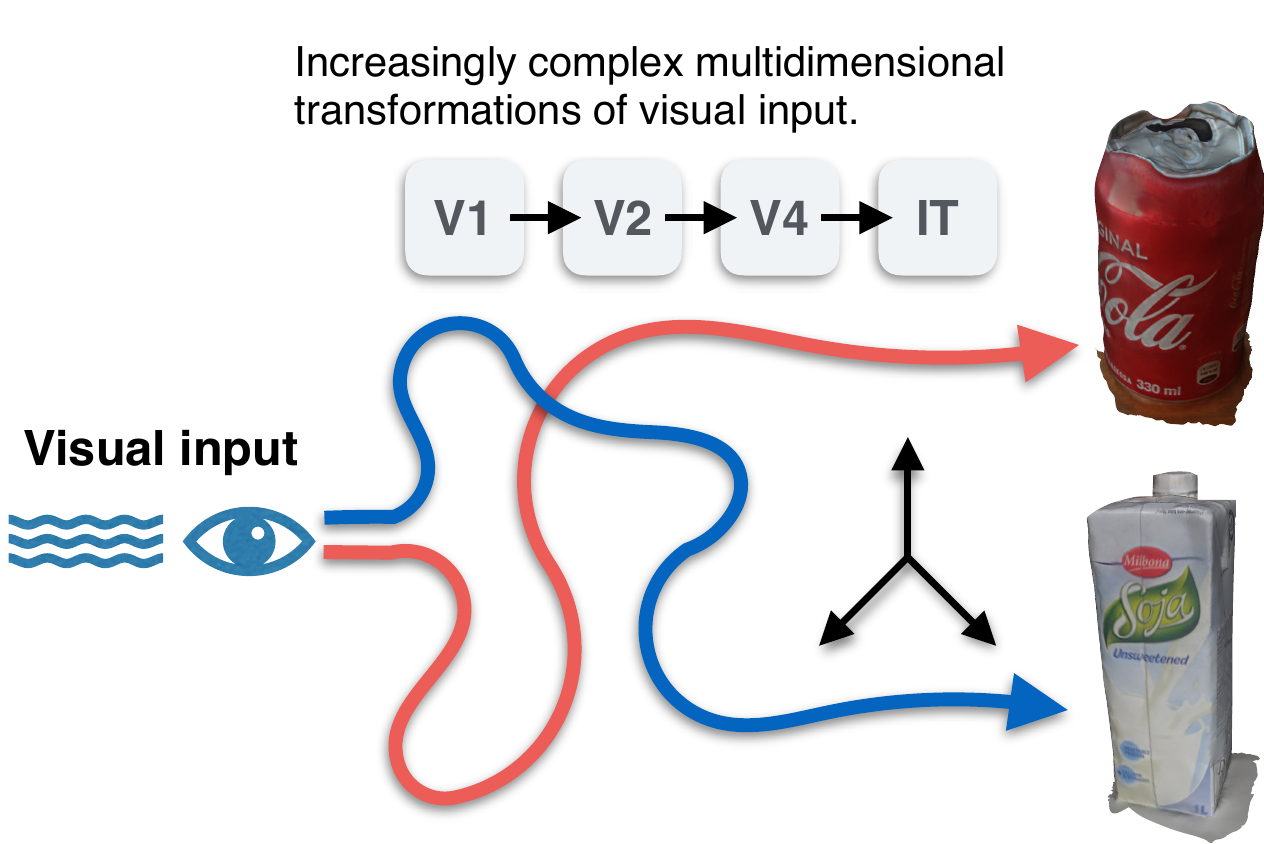}
\caption{\textbf{Hiearchial Transformations Models}. The visual input passes through the visual cortex, which performs complex multidimensional transformations making the transformed input increasingly invariant and specific. For basic categorization, the model in \cite{DiCarlo:2012em} suggest a final representation that is linearly separable as illustrated in the figure.}
\label{fig:visual_understanding_hiearchial_model}
\end{figure}

\subsubsection{Recent Ideas - Representational Transformations}

Recent ideas of visual understanding focus on tying together the neural
aspect of vision with higher-level cognition. Models of vision in
computational biology are similar in spirit to the deep neural networks
models used in machine learning. They characterize image recognition as
a computational feedforward hierarchal network ending in the inferior
temporal cortex (IT) which is generally considered the end of the
ventral pathway
\cite{Wallis:1997er, Riesenhuber:2000bw, DiCarlo:2007hs, DiCarlo:2012em, Isik:2016ts, Pinto:2008gj, Serre:2007kq, Serre:2007ir, Rolls:2012dt}.

The feedforward network continuously refines the retinal images through
linear and nonlinear transformations into complex multidimensional
representations that are increasingly invariant to the position on the
retina \cite{Riesenhuber:2000bw}
(fig.\ref{fig:visual_understanding_hiearchial_model}). The feedforward
network model is essentially a version of the multiple viewpoints model
which has replaced the stitching mechanism by transformations that
facilitate categorization and identification by separation in space.

The transformations of these models vary from model to model. In
\cite{DiCarlo:2012em} they are both high-dimensional and low-dimensional
projections correlating size wise with the neuronal populations of the
different cortices in the ventral pathway. The authors suggest that each
neuronal subpopulation tries to achieve a transformation, using
nonlinear functions such as logical-gates, max-pooling, and thresholding
of neuronal firing rates. These transformations flatten the object
manifold together with some form of learning rule that concentrates the
response to input areas where the object is usually found. The final
representation is an untangling of the object manifolds that make them
separable by a hyperplane \cite{DiCarlo:2007hs}. The model learns the
parameters in an unsupervised fashion from the temporal contiguity of
objects on the retina. We can find these construction principles in some
form or other in most hierarchical models e.g.~\cite{Rolls:2012dt}.

The descriptors produced by these models must account for the different
phenomena observed in human vision as outlined previously, as well as
the speed of human basic image categorization which is less than 200 ms
\cite{Riesenhuber:2000bw, DiCarlo:2012em}. One way of evaluating these
computational models of human vision is on standard computer vision
benchmark datasets. However, it has been argued that these datasets are
difficult to use for evaluating promising models since they have low
variation in viewpoint, position, size, etc., both for intra-class and
between-class categories, and contain contexts that covary with the
object category \cite{Pinto:2008gj, Rolls:2012dt}.

These models build on neurological evidence that recognition is not done
by matching to a single neuron but to a range of them. Neurons in IT
seems to encode features such as shape and other low-level properties of
the object rather than a complete representation \cite{Baldassi:2013gu}
and \cite{DiCarlo:2012em} notes that the weighted sum of intervals of
spiking patterns appears to be enough to explain object pose variations.

Additional findings in studies of primates \cite{Booth:1998tt}, show
that for familiar objects some of the neurons will encode viewpoint
invariant information, responding to all views of an object and that
most other neurons respond to specific viewpoints. The authors
hypothesize that the view-dependent neurons form associated clusters
that influence certain neurons to become view-invariant. These object
responsive neurons then allow for rapid object recognition for familiar
objects.

Recordings of IT neurons in monkeys also show that increasing
familiarity with an object correlates with an increase in neurons
encoding for the object \cite{Peissig:2007dt, Booth:1998tt}; implying
that familiarity refines the internal representation. Experiments on
primates have also shown the plasticity of neurons in the temporal
cortex when exposed to novel stimuli of the same category. In the
experiment, monkeys were first exposed to familiar faces 5-15 times
where the recorded neurons showed a stable response. When introduced to
novel faces the same recorded neurons altered their response to a
relative degree. This flexibility in altering the response suggests that
category encodings are continuously updated in a rapid fashion to
improve categorization \cite{Wallis:1997er}.

From a machine learning perspective, the idea of feature invariance and
multiple views are not a dichotomy. The untangling will by necessity
have to discard features that are not relevant. Multiple views of the
same or similar objects are actually necessary to find the most
efficient untangling. The untangled representations will have to be
close in space for it to be useful. From this perspective, the
normalization and interpolation of the multiple viewpoints model will
not be a problem since the untangling will automagically facilitate it.
It is important to note that the multiple viewpoints of an object are
not stored per se but it is \emph{the parameters of the network that are
the compressed storage form}.

The hierarchal nature of the networks also makes it possible to form
different representations for different tasks. For example, subordinate
categories can be represented higher up in the hierarchy since they need
fine-grained discrimination. It is also not inconceivable that the brain
uses these representations to build the prototypes and parts-based
models that we use as abstractions when reasoning about categories. For
a discussion on the advantages of feature hierarchy systems see
\cite{Rolls:2012dt}.

\subsection{Embodied Cognition - Categories \& Semantic Memory}

\label{subsubsec:embodied_cognition} An important part of object
understanding is the semantic memory, that is, facts, ideas, and
concepts that we can recall at an instance. In its most straightforward
form, this manifests itself as a deeper understanding of the meaning of
words, for example, objects and their properties. Storing of such
information does not come as discrete entities, as the number of
categories is infinite. The semantic memory is, therefore, best
characterized by parsimonious, flexible, intertwined, and shared feature
spaces of overlapping categories \cite{Martin:2001cp}.

A large body of evidence suggests that the brain stores properties of
objects close to where the functional unit for recognizing them are
\cite{Martin:2001cp, Martin:2007ck}. For example, color is stored close
to or in the area of the brain responsible for the perception of color
and knowledge about tool use is stored close to or in motor-related
regions. This means that category and object-specific information should
come from a weighted accumulation of information from the different
property-based regions \cite{Martin:2007ck}. This accumulation is
suggested to converge in high-level convergence zones that are far away
from the primary sensorimotor cortices \cite{Binder:2011wu}.

Experiments show that mentioning words denoting objects triggers
activity in areas of the brain where it first learned the properties of
the object. Memories can thus be triggered by reactivation of the neural
pattern elicited when learning about the object. Greatly simplified,
this implies that the concept of, for example, a cat triggers the sensor
modalities that involves the recognition of a cat. Further on, concepts
can have different meanings given the current context, implying that
they should be dependent on the convergence of different sensor
modalities.

Some researchers have in light of this proposed models of the brain as
performing mental simulations when cognizing
\cite{Gallese:2005fl, WBarsalou:1999ex}. This idea is backed up by
evidence of mirror neurons \cite{Rizzolatti:1996fo}. Mirror neurons are
neurons that activate when we think about a task, see or hear a word
denoting a task, and when we see someone else perform the task. For
example, seeing someone perform a grasping action invokes areas in the
brain of the perceiver associated with motor commands of grasping, even
mirroring the specific grasp that the person is performing.

Given this, and plenty more experimental evidence
\cite{Barsalou:2008ff}, it has been argued that to understand a concept
we must ground the meaning in sensorimotor input and that the actual
understanding of a concept is, in fact, sensorimotor simulation. This
idea is usually referred to as the embodied cognition hypothesis or
grounded cognition
\cite{Barsalou:2008ff, Barsalou:2016is, Gallese:2005fl, Martin:2001cp, Pulvermuller:2014jr},
where cognition according to this model manifests itself in form of
simulations, situated action, and bodily states \cite{Barsalou:2008ff}.
In essence, this is a model for solving the symbolic grounding problem
\cite{Harnad:1999uv}.

\cite{Kiefer:2012bi, Pulvermuller:2014jr} argues that concepts, as
physically realized in the brain, are distributed representations
connected via action-perception circuits (APC) that links the necessary
sensorimotor modalities and higher convergence areas. The APCs are
learned through exploration by finding correlates in congruent
activation of sensor modalities.

Critics of grounded cognition come from the traditional cognitivist
viewpoint. They consider concepts as amodal, that is, abstract symbols
separate from the sensorimotor system. We can think of the separation as
a cognizing unit manipulating symbols for planning and a sensorimotor
system infusing the symbols with meaning. For example,
\cite{Mahon:2008fv} argues that sensorimotor activation can be
epiphenomenal, a byproduct of symbolic manipulation. Patients with
lesions in motoric regions can for example reason about the concept of a
tool even though they are not able to use it. As such, the amodal
viewpoint is that the abstract concept triggers the sensorimotor
simulations.

Nonetheless, studies of patients with various types of motor damage show
impairment in comprehending action words such as tool use.
\cite{Binder:2011wu} suggest that the sensorimotor activations are
hierarchical, modular, situational, and frequency dependent. We can then
explain the double
dissociation\footnote{If damage $X$ to the brain affects functionality $A$ but not functionality $B$ and damage $Y$ does not affect functionality $A$ but affects functionality $B$ then the parts of the brain associated with these functionalities are considered to be double dissociated. Double dissociation is stronger evidence of the independence of location in the brain of these different functionalities.}
given above, that is often used as an argument against embodied
cognition, as a degradation in the functionality of the structure. For a
recent discussion on the topic of amodal and embodied cognition see
\cite{Barsalou:2016is}.

\subsubsection{The Dorsal Pathway}

The dorsal pathway processes visual input for action, parsing objects
and scenes in a person-centered reference frame. Research on vision has
mainly concerned itself with cognition and not how vision guides action
\cite{Goodale:2011ff, Grafton:2010jo}. Understanding the processing of
vision for action is therefore not as developed as that of vision for
cognition.

Current understanding of the dorsal stream is similar to the model of
the ventral stream as a series of transformations of the visual input to
representations that facilitate action planning and execution. Some have
even suggested that features in the dorsal pathway are computed in a
feedforward hierarchy similar to the ventral pathway
\cite{Rolls:2007gu}. The visual inputs flow from V1 to V2, and on to V3,
that projects to areas involved in action planning and execution. The
visual flow is illustrated in fig.\ref{fig:brain_image_proceesing}

More recent models of the dorsal pathway suggest that it gives rise to
three distinct pathways supporting spatial working memory, visually
guided action, and spatial navigation \cite{Kravitz:2011vy}. The three
pathways, parieto-prefrontal, parieto-premotor, and parieto-medial
temporal, all found in the posterior parietal cortex integrates
information from the central and peripheral visual fields, forming
reference frames relative to the body and to the eyes. The pathways
connect a range of functions such as spatial working memory, the
top-down guidance of eye-moment, optical flow, depth information, world
and object space reference frames, navigation, and the reaching and
grasping of objects \cite{Kravitz:2011vy}.

The dorsal stream is generally not thought to encode for objects in a
view-invariant manner as in the ventral stream. However, see
\cite{Farivar:2009tt} for a review of the evidence of viewpoint
invariance in the dorsal stream, as well as a discussion on different
cues processed in the two streams and their integration in object
perception and understanding. Viewpoints in the dorsal stream are
instead interpreted as different objects \cite{James:2002gf}. They are
not parsed for global contextual interpretation as in the ventral stream
but provide an encoding that favors ease of visuomotor transformations
for tracking, reaching, and grasping \cite{Chinellato:2015uj}. This
sensitivity to viewpoint makes sense since grasping is an object
centered action. The dynamics and the local shape of the object are
central to physically performing the reaching and grasping action.
However, as we shall see contextual information plays a big role in
planning and choosing which motor commands to perform.

\subsubsection{Ventral Stream Influence on Grasp Planning}

\label{chp1:cognition_grasping:ventral_stream_grasping} As outlined
above, the dorsal processing concerns itself with visuospatial
transformations used for actions such as grasping, disregarding semantic
knowledge. The ventral pathway, on the other hand, is thought to only
influence motor planning indirectly.

Experiments with the patient D.F. showed, for example, inability to
infer properties of size, shape, and orientation verbally or manually
but showed normal pre-shaping and rotation of the hand when grasping
objects \cite{James2463}. This and other studies have been taken as
evidence for the separation into the two different representations
catering to action and conceptual understanding.

The observations from D.F. are not all consistent with the idea of
perception-action separation. For example, D.F. did not use visual cues
when modulating grasp force something which would have had to come from
the ventral pathway as this should be highly dependent on the object and
its material \cite{Schenk:2010jq}. Further on, D.F's ability to grasp
objects correctly are reliant on a specific set of visual cues, and when
those cues were perturbed she would fail more frequently than healthy
subjects. In \cite{Schenk:2010jq} the authors argue that this is
evidence that the ventral stream processes additional depth and distance
cues which are then used in the grasping process. This is echoed in a
review by \cite{Neri:2004ix}. There the authors suggest that the dorsal
pathway encodes primarily for absolute disparity while the ventral
system is concerned with relative disparity. \cite{Schenk:2010gm} argues
further that observations of the dorsal and ventral streams should be
built on the premise of integration instead of separation. They argue
that D.F.'s degradation should be seen as if she lost specific visual
cues that are reliable under certain circumstances and tasks. As such
she would have to rely on the accumulation of less reliable cues to
guide recognition and action.

Yet in recent experiments on binocular disparity in macaque monkeys, it
is concluded that relative disparity exists in both visual streams,
although serving different purposes \cite{Krug:2011fg}. The authors
propose that the ventral stream uses relative disparity to judge the
shape of objects. This would explain D.F.'s troubles in discerning
spatially adjacent surfaces since relative disparity is more accurate
than absolute \cite{10.1371/journal.pone.0012608}. The dorsal stream
uses relative disparity to aid in the segmentation of moving features.
This implies that interaction with static objects relies on ventral
stream information for more delicate manipulation.

There is further evidence of dorsal-ventral cross-talk. An fMRI study
\cite{Amedi:2001dc} found somatosensory activation in parts of the
ventral pathway implying that tactile information might be coupled with
visual object knowledge. Others have found that recognition of tool use
integrates information both from object attributes stored in the ventral
stream and from motor-based properties in the dorsal stream
\cite{Chao:2000bf} in accordance with the idea that memory of categories
and attributes are stored close to the area for detection
\cite{Martin:2001cp}. Areas close to the AIP, the area that is
responsible for transforming visuospatial features into hand
configurations, are also active during action and objection recognition
which means they might provide information about functional properties
of the object. Further on, a recent analysis of fMRI data found that the
ventral pathway responds to weight or textural density as part of the
visual processing for grasping \cite{Gallivan:2017jz}.

For reviews and discussions on the ventral and dorsal separation see
\cite{Goodale:2011ff, Schenk:2010jq, Schenk:2010gm}.

\subsection{Summary}

The prevalent model of human vision is that of the two pathways, one for
conceptual understanding answering \emph{what} it is we see, and one for
action answering \emph{how} we should do what we want to do. One should
not view the separation as absolutely distinct as there is plenty of
evidence of crosstalk that helps solve the how or what questions,
\cite{Goodale:2011ff, Schenk:2010jq, Schenk:2010gm}.

The ventral pathway performs basic level categorization. It is modeled
as a feedforward network where representations along the network are
refined along the cortical visual areas, becoming more invariant to
scale, luminance, position, etc. The feedforward models are based on a
schema of neuronal functionality and the idea of temporal firing rates
as representations. Despite their intuitive and simple form, the basic
feedforward networks only model a fraction of visual understanding.
Visual cognition requires constant recalibration in the form of
attention and refinement implying that a complete model of vision will
have to take into account the existing feedback projections along the
ventral pathway. It will also have to go beyond basic recognition
explaining abstraction, deeper visual understanding, etc. There is
plenty of arguments against framing visual understanding as solely
consisting of object recognition \cite{Cox:2014km}.

Further on, converging evidence points to that humans ground concepts in
sensorimotor input in some form or other, as opposed to being amodal.
The grounded concepts bridges object properties with possible
affordances and motor programs that perform the affordance.

The dorsal pathway handles spatial understanding for interacting with
objects. It transforms visual input into representations for planning
and execution of actions in the motor cortex. Representations in the
dorsal pathway are in general not view-invariant and the dorsal pathway
does not store representations of graspable objects for long, most
probably due to the temporal aspect of actions.

There is plenty of evidence of cross-talk between the two streams where
representations in the ventral pathway are thought to influence, for
example, grasp planning and other activities reliant on semantic
knowledge.

\section{Human Grasping}

\label{cognition_grasping:human_grasping} Neuroscience has yet to solve,
on a broader level, how perceptual inputs guide actions
\cite{Castiello:2007iy}. There is a large body of work on prehension in
monkeys \cite{Rizzolatti1988}, however, the human physiology and
everyday actions are somewhat different from monkeys, and the putative
homologs in humans are still not fully understood
\cite{Castiello:2005gq}. We give a simplified outline of the visual
processing for grasping in primates in
fig.\ref{fig:brain_image_proceesing}. For an in-depth exposition of the
areas in the brain, that are active during grasping, and the flow of
information see \cite{Chinellato:2015uj}.

The visual information in grasping comes majorly from the dorsal stream.
V3 projects into the intraparietal sulcus (IPS). Experiments on macaques
show that the IPS processes visual stream for action-perception
coordination \cite{Grefkes:2005gf}, that is, arm and eye movements. The
IPS contains the caudal intraparietal sulcus (CIP) which process 3D
features, axis, and surface orientation, from the information in V3, in
a view-centered manner \cite{Chinellato:2015uj}.

CIP connects to the anterior intraparietal area (AIP) which is thought
to be central to grasp planning. Neurons in the AIP discharge during
object manipulation, object observation, and even to be sensitive to
manipulation during dark and light luminance conditions.
\cite{Culham:2004uy} found that AIP neurons, in monkeys, respond
preferentially to specific manipulations of specific objects in addition
to being selective for shape, size, and orientation.

The AIP is thus thought to perform visuomotor transformations of
forwarded visuospatial object features. It transforms these features,
such as surface orientation in depth and texture gradients into grasp
plans which it then forwards to the motor cortex for execution. The AIP
is also active during the grasping process where the evidence points to
it being critical for monitoring and recalibrating grasp movements
\cite{Chinellato:2015uj}.

The AIP connects to the premotor cortex. Part of the premotor cortex is
the ventral premotor cortex (PMv) which contains area F5. F5 fires
during specific object-related hand movements and in the presence of a
3D object \cite{Grefkes:2005gf}. Recordings also show some F5 neurons
discharging for end-state goals \cite{Rochat:2010hy}. Experiments in
\cite{Rizzolatti1988} found sets of neurons discharge in these regions
during specific types of prehensile movements suggesting that these
combinations of neurons are a motor vocabulary for elementary motor
acts. The first discovery of mirror neurons was in F5. Experiments
showed activation in F5 when a Macaques was observing other actors
perform grasp actions.

PMv neurons in monkeys are selective for the type of required prehension
for grasping an object, that is, the grasping action as a whole not
controlling for the individual digits, together with the applied force.
Experiments on monkeys have in fact found a striking accuracy, 89\%, in
predicting grasping actions based on the activity in the premotor cortex
\cite{Grafton:2010jo}, see also \cite{Fabbri:2016bd}.

The PMv receives information from the AIP and is thought to output a
representation activating motor programs of the object affordance which
is then combined with other visual cues to configure and orient the
hand. This information is then forwarded to the primary motor cortex
(M1) for execution \cite{Castiello:2007iy, Grafton:2010jo}.

fMRI studies also suggest that the human homolog of the PMv integrates
additional modalities such as somatosensory \cite{Ehrsson:2000vo}. The
authors suggest that there are different areas responsible for precision
and power grips, where precision grips might use somatosensory
information from the posterior parietal cortex. The use of more tactile
information in precision grips is sensible. They tend to be complex and
followed by in-hand manipulation which is much more reliant on tactile
feedback. In addition, \cite{Ehrsson:2000vo} also found that precision
and power grips activate different parts of the human cortex. The
precision grip activated a larger combination of cortical areas
especially the PMv while power grips associated more with M1. This seems
to be in accordance with the notion that precision grasps require more
planning and coordination than power grips.

\begin{figure*}
    \includegraphics[width=\textwidth]{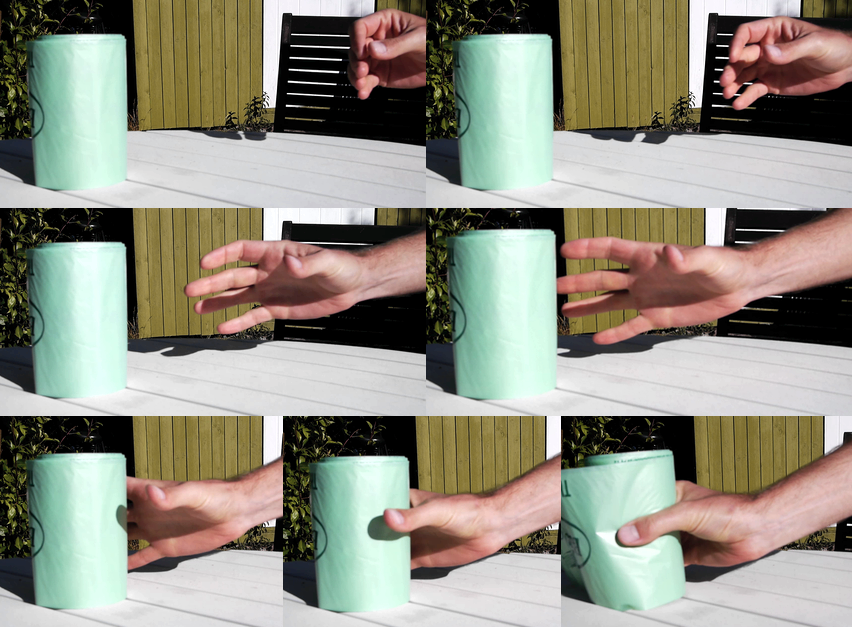}
    \caption{\label{fig:grasp_sequence_example} A sequence of snapshots of the reaching movement for grasping a cylinder-shaped object. Notice how the pre-shape develops, with the grasp aperture increasing during the reaching movement, correlating with the shape of the object, and then decreases as the hand reaches the object. We also note that the actor places the grasp at the mid-section on the object as to get a stable grasp for picking up the object as opposed to if the actor intended to give the object to someone where a top-grasp would be more convenient.}
\end{figure*}

\subsection{Grasp Planning and Execution}

Human grasping can be characterized by a set of phases
\cite{MacKenzie:1994wh}.We illustrate the grasping movement in
fig.\ref{fig:grasp_sequence_example}.

\emph{The first phase} is concerned with finding a suitable plan of
execution. In general, this means choosing a grasp position on the
object, pairing it with an initial estimate of the hand configuration
that will fit the local geometry of the object, and then adding the
proper motor commands that involve the reaching and grasping movement.

\emph{The second phase} involves executing the motor commands for the
reaching movement and pre-shaping of the hand to comply with the local
geometry of the object at the chosen grasp position. During this phase,
there is a continuous recalibration of the trajectory and the hand
configuration, where the hand reaches a maximum aperture around 60-70\%
through the reaching movement correlating with the size of the object
\cite{Castiello:2005gq}.

\emph{The third phase} is during and at contact with the object where
hand reconfigurations are made to stabilize and maintain the grasp. This
phase is dominated by the aggregation of sensor modalities such as
tactile, visual, auditory, and proprioceptive working together to give
feedback about the stability of the grasp and how that stability is
maintained.

We will concern ourselves with the initial phase, the planning, as this
articles' main focus is on how the visual understanding influences the
grasp planning process.

\subsection{Effects of Task and Object Properties on Grasp Planning}

\label{subsec:task_object} 
\begin{figure*}
    \centering
    \includegraphics[width=0.75\textwidth]{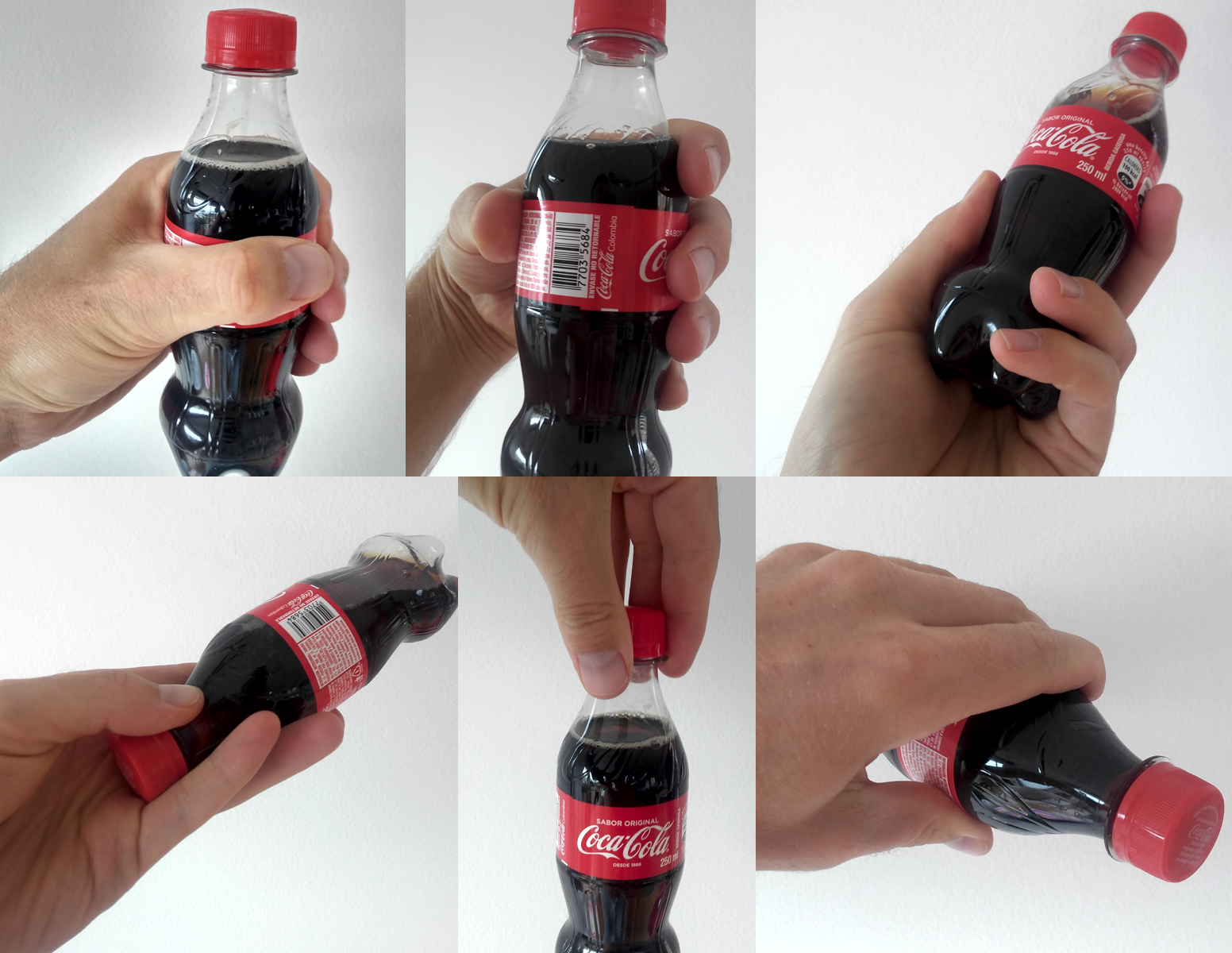}
    \caption{\label{fig:grasp_examples} Task-specific grasps for the tasks: drinking, drinking, giving, giving, picking up, pouring. Clearly, task is the major factor in deciding approach vector, grip position, and grip configuration. Notice also how the pinky finger in the first two images acts as stabilizing factor by being situtated underneath the curvature of the bottle. A grasp higher up without the pinky finger stabilitziation will be much more unstable. We mostly likely learn this type of stabilization by repeated interaction implying that this type of grasp is to some extent reliant on semantic memory.}
\end{figure*}

Many factors affect the planning of a grasp.
The post-grasp task, that is, what one intends to do with the object, is
central to planning with respect to the reaching movement
\cite{Ansuini:2006eu, Ekvall:2005ue}, hand configuration
\cite{Napier:1956tn, Rosenbaum:2012ki}, placement
\cite{Jeannerod:1999ke}, and the applied load and prehensile force upon
contact \cite{Flanagan:2009ux}. In addition, the properties of the
object also affect grasp planning, placement, and load forces. Apart
from the shape of the object, estimations of dynamics, weight, and
texture are the predominant factors in planning, through slow-changing
priors that guide the initial plans. We give a number task-specific
grasps in fig.\ref{fig:grasp_examples} to illustrate how task might
affect grasp placement.

\subsection{Task \& Semantic Memory}

\label{subsec:task_semantic_memory} Human grasp planning is seemingly
preemptive in that it continuously updates and computes potential
grasping actions and affordances \cite{Belardinelli:2016ib}. Before
initializing the grasping action the eyes typically focus its attention
on parts of the scene that it anticipates being the focus of an action.
Experiments show that the eyes fixate on a set of landmarks that are
central to the task \cite{Johansson6917}. These landmarks are most
likely used to anticipate potential grasp configurations and actions.

Once set on a target object for grasping, experiments show that the eye
fixations show a preference for regions where the brain intends to place
the index finger of the grasping hand indicating hand-eye coordination
in planning \cite{Belardinelli:2016ib}. \cite{Belardinelli:2016ib} even
suggest that it might be possible to predict which object manipulation
is about to take place given the current fixation point and object.

This anticipatory behavior also manifests itself as subconscious
computations of potential actions for objects found in the vicinity. In
a series of experiments, \cite{Tucker:2001ey} showed that humans using
different grasp types to signal the category of objects had faster
response time with object compatible grasps. This led the authors to
suggest that the brain also represent objects in general motor responses
that are potentiated irrespective of the agent's intentions, that is,
the brain is subconsciously computing potential action possibilities.

A later fMRI study \cite{Grezes:2003bz} showed grasp types compatible
with the object indeed activated the parietal, dorsal premotor, and
inferior frontal cortex. Another fMRI study by \cite{Chao:2000bf}, where
participants named the categories of pictures depicting objects, also
showed activation in ventral premotor areas, specifically for pictures
of tools. These motor response representations led the authors of
\cite{Ellis:2000uv, Tucker:2001ey} to coin the term
\emph{micro-affordances} to denote possible grasping actions not
necessarily involved in just one type of affordance.

Planning of the grasp also considers the post-grasp dynamics.
Experiments show that at least for simpler tasks \cite{Hermens:2014jl},
the chosen grasp is part of a broader strategy for maximizing control at
the end of the task sequence, a so-called ``end-state comfort''. This
effect has been shown in a range of works from Rosenbaum and others
\cite{Rosenbaum:2012ki, Herbort:2014dj}.

The initial realization came from observing a waiter filling glasses
with water. The waiter grasped an upside-down standing glass with the
thumb in the direction of the opening of the glass, rotated it, filled
it with water, and put it down, thumb facing up. Rosenbaum noted, that
the waiter chose the initial awkward posture for a less awkward
post-grasp posture. After initial experiments, the optimization strategy
was named the end-state comfort effect. Further analysis, however,
indicated that humans choose grasps that optimize the control exerted
over the grasped object when it is most needed. Additional experiments
involving handing-over of objects showed that humans choose awkward
postures to enable the receiver to perform an action with the object.
The actor thus actively takes another actor's intentions into account.

Task also affect the speed of the reaching movement and hand shaping
\cite{JohnsonFrey:2004ek, Ansuini:2006eu}. Experiments involving
grasping the same object for different tasks \cite{Ansuini:2006eu}
showed that when the end task required greater control the pre-shaping
was more gradual and the reaching movement of the hand was much slower.
In the simpler end task, the subjects formed the hand shape directly to
comply with the place for a grasp while in the complex task, the
pre-shaping happened gradually during the reaching movement. In sequent
experiments, \cite{Ansuini:2007fn} showed that using the same object but
for different actions affected reaching duration such that it was
markedly slower if no task followed the grasp. These results suggest,
again, that the grasp planning anticipates the end-state. The authors
explain the longer reaching time for the no-task grasp as relying more
on tactile feedback as opposed to when there is a post-grasp task that
requires planning and taking dynamic constraints into account.

Grasp placement for more complex tasks also appears to involve semantic
memory. \cite{Creem:2001ji} showed in experiments, involving objects
with handles rotated in different orientations, that when the subjects
were simultaneously involved in a memory retrieval task they tended to
pick up the object by the nearest point regardless of the distance to
the handle. However, when the memory retrieval task only involved
spatial or verbal working memory components, the subjects the grasps
landed on the handles with higher likelihood regardless of orientation.

These findings indicate the importance of memory retrieval for
performing more complex grasping procedures. It also implies that humans
perform the simple act of reaching and grasping automatically without
the involvement of semantic memory. Additional evidence of the
involvement of memory in the grasping process comes from
\cite{McIntosh:2008jd}. The authors showed that objects of familiar size
modified both reaching and hand amplitude. This effect was even greater
when they removed binocular cues implying an increase in reliance on
memory cues.

Representations in the brain of objects are also affected by task
\cite{Harel:2014cp, Bugatus:2017wl, VaziriPashkam:2017cu}.
\cite{Harel:2014cp} performed an fMRI study of twenty-five subjects that
carried out six different tasks that required judging physical
properties: fixation, color, tilt, or conceptual properties: content,
movement, size. First, they showed the subjects an object and then
tasked them with judging a property of the object. The study showed that
the task context strongly affected the response in the ventral pathway
such that it was easier to decode which object the subjects judged
within a task compared to across-task. This was also taken as further
evidence of the top-down modulation of visual processing as the task
affected object processing.

Similar experiments in \cite{VaziriPashkam:2017cu} showed that task
affected processing in the dorsal stream while the ventral stream was
less affected and where the early visual regions showed a higher
encoding for category than task. Further on, the experiments also showed
that the filtering out of salient task-irrelevant features was greater
in the dorsal pathway indicating the stream's relevance for parsing
visual input into action.

% \begin{figure}
%     \small
%     \begin{subfigure}[t]{.15\textwidth}
%         \includegraphics[height=3cm]{imgs/grasp_drinking.png} \label{fig::trTool}
%     \end{subfigure}
%     \begin{subfigure}[t]{.15\textwidth}
%         \includegraphics[height=3cm]{imgs/grasp_handing_over_1.png} \label{fig::trTool}
%     \end{subfigure}     
%     \begin{subfigure}[t]{.15\textwidth}
%         \includegraphics[height=3cm]{imgs/grasp_handing_over_2.png} \label{fig::trTool}
%     \end{subfigure}   
%     \begin{subfigure}[t]{.15\textwidth}
%         \includegraphics[height=3cm]{imgs/grasp_holding.png} \label{fig::trTool}
%     \end{subfigure}           
%     \begin{subfigure}[t]{.15\textwidth}
%         \includegraphics[height=3cm]{imgs/grasp_picking_up.png} \label{fig::trTool}
%     \end{subfigure}    
%     \begin{subfigure}[t]{.15\textwidth}
%         \includegraphics[height=3cm]{imgs/grasp_pouring.png} \label{fig::trTool}
%     \end{subfigure}   
%     \caption{\label{fig:grasp_examples} \small }
% \end{figure}

\subsection{Shape}

\label{subsec:shape} The shape of the object affects grasp placement by
determining the center of gravity as well as offering places for
support. For example, in fig.~\ref{fig:grasp_examples} it is easy to see
how the pinky finger uses the curvature of the bottle as a stabilizing
factor. Experiments in \cite{10.1371/journal.pone.0025203} on two
differently shaped plastic water bottles involving the task of pouring
and moving exposed effects of task and object properties on the grasp
performance. The bottles used in the experiment consisted of one
ordinary cylindrical shaped bottle and one bottle with concave
constriction similar to the bottle in fig.\ref{fig:grasp_examples}. The
bottles were either half-full or full.

In the experiments, the subjects grasped the full bottles higher up than
the half-full and more internally towards the center of gravity. The
same effect was observed for the pouring actions versus the moving
action. The grasps made for pouring were placed higher up and more
towards the center of gravity. The shape affected the grip apertures by
generating smaller apertures for the moving action of the full concave
bottle. Task also affected the movement time where the moving action was
faster than pouring.

These results are evidence of the anticipatory nature of grasping. The
grasp planning phase clearly took into consideration both the shape, in
terms of the digit placement, and the estimated weight of the bottles,
to determine how they would affect the post-grasp task. The authors
suggested that the reason for the moving action eliciting a faster
reaching movement is likely due to it requiring less precision and
calibration.

Similar experiments in \cite{Ansuini:2006eu}, involving placing an
object inside differently sized niches, also showed this effect. The
reaching movements made for niches that required less precision were
faster and the pre-shaping of the hand reached its final shape almost
immediately opposed to the grasps for niches that required more
precision. This is in line with the fMRI results \cite{Ehrsson:2000vo}
described earlier which showed that precision grips generate larger
neuronal activity compared to power grips. This is sensible as demanding
post-grasp tasks are more difficult to plan and are likely to require
more recalibration during the reaching movement.

Additional evidence of how task affects digit placement comes from
experiments in \cite{Craje:2011eg}. The experiments involved pouring and
lifting a common glass juice bottle and showed that the task and the
weight of the bottle significantly affected digit placement. For
example, the pouring task consistently showed greater distance between
thumb and index finger, something the authors suggest was to facilitate
the rotating movement of the bottle. We have illustrated a similar grasp
in fig.\ref{fig:grasp_examples} where the two first images indicate the
stabilization factor by the pinky finger. Further on, the first and last
images show the difference in distance between thumb and index finger
digits for a power grasp and a precision grasp.

Shape is also a factor when computing the applied grip force.
\cite{Jenmalm:2000ui} showed in a set of experiments of precision grips
on an object, which changed curvature unpredictably between trials, that
humans use the curvature and kinematics estimated by vision to predict
the required movements and grip forces that produce a stable grasp. The
experiments also showed that anesthesia had little impact on the
adaptation of the grip force indicating that visual cues are the
predominant variable in grip force estimation. However, vision was of
little help to the anesthetized subjects in modulating the grip forces
to balance friction. This implies that past the contact phase tactile
feedback is the biggest factor in modulating the grip force.

\subsection{Size and Material}

\label{subsec:size_material} A classic experiment by primary school
physics teachers is to show two objects of different size and material,
and ask which falls the fastest. Whereupon most students, not familiar
or sometimes even familiar with the laws of gravity, answers the larger
or by material seemingly heavier one. Priors on object weight also play
an important role in computing load forces needed for lifting and
performing actions with objects.

Humans base their prior on the size and material of the object.
Experiments involving objects whose surface displays a material
different from the interior showed that subjects misjudged weight and
applied erroneous load forces. \cite{Baugh1262, Buckingham:2009dd}
tasked humans with lifting cubes of varying sizes, and with different
surface and interior materials (brass and wood in \cite{Baugh1262}, and
metal, wood, and expanded polystyrene in \cite{Buckingham:2009dd}). The
experiments showed that the subjects frequently misjudged object weight
but learned to adapt the load force after a couple of additional
interactions.

Interestingly, the prior on object weight in these experiments was
seemingly stable and did not update with new information. The subjects
consistently made errors when judging object weights even after
receiving feedback and being asked about the object weight again. The
applied load force, on the other hand, started to adapt after the first
trial. The authors suggest that humans rely on two distinct
representations for estimating weights. A slow-changing
material-density-volume prior that we use for initial estimates of
object weight. When the prior proves to be wrong the estimates tend to
shift and rely to a greater extent on a combination of priors and
sensorimotor memory.

A recent analysis of fMRI data supports the idea of these types of
priors. It showed the ventral pathway responding to weight or textural
density as part of the visual processing for grasping, and that these
associations, in fact, are learnable \cite{Gallivan:2017jz}.

Size, material, and dynamic priors are especially valuable when placing
precision grips. Precisions grips require complex interaction between
different load forces. They must predict both weight and friction
between the fingertips and the object to prevent slippage as well as
predict the dynamic behavior of the object under load forces. These
types of priors have led to suggestions that humans form internal models
of object dynamics.
\footnote{See \cite{DBLP:journals/corr/AgrawalNAML16} for a deep learning approach to learning robots to develop similar models.}.
Sometimes with enough training, the dynamic model even becomes specific
to a single object \cite{Flanagan:2009ux}.

Many of the approaches for stability prediction in robotic grasping have
usually assumed knowledge of friction constants etc. These approaches
are obviously not tenable and the trend is towards data-driven methods
e.g.~\cite{DBLP:journals/trob/BekirogluLJKK11, 6943027, DBLP:journals/corr/AgrawalNAML16, DBLP:conf/iccv/RogezSR15}.
However, the lack of reliability and resolution in robotic tactile
sensors compared to humans leaves only so much room for improvement in
the sophistication of the control and learning algorithms.

\subsection{Heuristics}

\label{Human-Grasping:Heuristics} Experimental evidence shows that
humans view objects in a holistic manner but when grasping they ignore
features that are not pertinent to placing the grasp
\cite{Goodale:2011ff}. These results suggest that there are heuristics
involved in grasp planning and that we can explain them in specific
features of the object.

A study \cite{Feix:2014cy, Feix:2014bw} of four professionals manual
laborers analyzed 7770 instances of object-task grasps of roughly 306
objects for 231 tasks collected during an 8-hour window. The analysis of
a decision tree classifier fitted to the data showed \(47\%\)
classification accuracy for grasp type for the attributes: dimension,
mass, roundness, functional class, task constraint, grasped dimension,
force, and rigidity. The most discriminative of the recorded features in
predicting grasp type were: object dimension, task constraints - the
degrees of freedom in rotating and translating the object, and the mass
of the object.

That object dimension is a good discriminator comes as no surprise. Most
tools are, for example, elongated and usually has a handle that requires
a specific grasp. Analysis of the data, in fact, showed that the
subjects had a clear tendency to grasp objects along the smallest
dimensions of the object. The constraints features are also logical as
manual labor involves repetitive tasks with specific objects with
specific grasps.

Another interesting aspect of the study is that the combination of
object and task constraints increased the classification accuracy
significantly compared to either feature alone. This is clear evidence
that categorization of required grasps needs to include both object
features and features involved in the action. It is also evidence of the
significant involvement and interplay of action with the object category
and how the affordance of an object shapes how we categorize and cognize
about it.

In another recent study \cite{Fabbri:2016bd}, participants grasped
objects that varied in size, elongation, and shape, using grasps with
different combinations of digits plus passively viewing the objects.
fMRI recordings of the subjects showed that the feature that was coded
strongest for in the different brain regions was elongation followed by
shape and size. The authors suggest that this reflects the importance of
dimension in selecting grip configuration and wrist orientation. In
addition, the results showed that the number of digits used in a grasp
was a better model for explaining the activations than the type of
precision grasp.

The above results show that simple features of an object might be enough
to derive simple heuristics for classification and grasp planning that
work well in a majority of cases. In the subsequent chapters, we will
explore this idea by formulating a stack of simple features and try to
learn from the data which feature works best.

\subsection{Summary}

We can decompose grasping into a set of phases summarized as planning,
reaching, and contact. The planning phase is preemptive in that the mind
consistently is anticipating and computing action possibilities, and
their associated motor programs for objects in the surrounding. The
planning is also preemptive in that it tries to predict the end-state of
the task such that the grasp maximizes control over the object.

Apart from the task, placement, and force in grasp planning depend on a
number of properties of the object. The properties are first and
foremost shape, material, and size. Humans learn slow-changing priors of
how materials respond to interaction. The elongation and size of an
object are good predictors for how and where a human will place
affordance-based grasps implying possible underlying heuristics for
placing grasps. On the whole, the accumulated evidence suggests a
complex interaction between factors of weight, material, shape, and
task. Further on, it shows that humans are efficient at exploiting and
estimating the involved factors to plan stable grasps that anticipate
future manipulation.

\section{Discussion And Conclusion}

It has been argued that,

\begin{quote}
vision began not as a system for perceiving the world, but as a system
for the distal control of movement \cite{Goodale:2011ff}
\end{quote}

This idea contrasts strikingly with the research done in computer vision
which has focused mainly on answering the \emph{what} question. The main
reason for this is most likely the effort involved in obtaining labeled
data. Answering \emph{what} is also less ambiguous than answering
\emph{how} in that categorization is a binary question while \emph{how}
spans complex interactions of body, object, and outcomes. It may also be
that \emph{what} is something that involves everyday conscious decisions
and often is explainable in the rule-based counting of features,
meanwhile, \emph{how} is subconscious and thus more complicated to
explain explicitly. This shows itself in that learning of actions are
primarily done by imitation and sensorimotor exploration.

The focus on answering \emph{what} has influenced the creation of many
vision-based grasping approaches in form of feature representation and
by forcing the problem into a form suitable for vision-based
discrimination,
e.g.~\cite{Saxena:2008wn, Lenz:2013uz, 7759657, Redmon:2015wj}. The
division of labor in human vision is a strong indication that robots
should use different features for action and recognition. Features for
grasping, for example, needs to be less focused on saliency and more
focused on shape and material, and how the visual understanding of shape
and material relates to the gripper configuration and force space. Data
for learning features should come from interactions with shape and
materials and will require more exploratory approaches as well as
learning from demonstration.

This is already being explored to some extent. In
\cite{DBLP:journals/corr/LevinePKQ16} the authors let a set of robots
perform 800 000 grasp attempts and uses the data to learn hand-eye
coordination for monocular images using reinforcement learning with a
convolutional neural network. \cite{DBLP:journals/corr/AgrawalNAML16}
hypothesize that humans have an internal physics model that allows them
to understand and predict how an action will affect an object and use a
siamese CNN to model a similar understanding. And in
\cite{DBLP:journals/corr/PintoGHPG16} the authors start with a base-net
that branches out into nets specialized for grasping, pushing, and
pulling actions. The base-net provides the basic processing of the
visual input and then receives feedback from the specialized nets on how
to improve. The approach thus mimics the human visual processing in the
idea of general preprocessing and then specialization. \cite{Gao:2016vs}
explores haptic adjectives such as compressible or smooth from the
fusion of visual and haptic data. For a thorough review on interactive
perception and further arguments for actionbased perception see
\cite{DBLP:journals/trob/BohgHSBKSS17}.

We also saw that human grasping is highly dependent on context and task.
Research has shown that the brain generates different representations of
an object depending on which task or context we view it in. Robotic
grasping research, on the other hand, has traditionally focused on
analytical measures of grasp stability disregarding context and task
e.g.~\cite{Bicchi:2000wr}. With an increased availability of grasping
data and tools for 3D vision, research has moved towards data-driven
grasping approaches. These methods rely in general on matching features
on the object to good hand configurations where they measure good
according to heuristics, matching to stored grasp-feature relations, or
minimizing a machine learning loss function \cite{6672028}.

Even though these data-driven approaches have been successful they have
failed to broach the broader subject of how object understanding and
intention affects how an agent should and can manipulate objects. Grasp
synthesis algorithms can heavily reduce the infinite number of grasping
positions on an object that it needs to consider by taking task into
account; as there is a limited set of positions for grasping an object
to complete a task successfully. Humans utilize this strategy of
optimizing post-grasp control sometimes placing an initial awkward grasp
to optimize for the end-goal.

A handful of efforts have incorporated task in robotic grasping.
\cite{Song:2010gh} trained a Bayesian Network (BN) relating object
properties with task, grasp, and constraint features. From the BN they
could produce probabilistic maps of hand pose over the object
conditioned on the task and object properties. \cite{Nikandrova:2015uu}
formulated a probabilistic model over task, stability, and known object
models to find stable grasps. \cite{Hjelm:2014uc, detry2017b}
modularized grasping into two modules one focused on matching known
grips to local properties on the object while the other computed the
task probability for a gripper position given known task-specific
grasps. The modularization enabled the transfer of grasps and task
constraints to novel object task combinations.
\cite{DBLP:conf/corl/AntonovaKSK18, DBLP:conf/humanoids/KokicSHK17}
leveraged deep learning to generalize task-constraints given by
pixel-wise ground-truths denoting affordance bearing parts.

In addition to task-based priors, we saw that humans use priors based on
material, size, and shape together with sensorimotor memory adaptation
when priors are wrong. To the best of our knowledge, only
\cite{Hjelm:2015hw} has approached the problem of learning priors.
\cite{Hjelm:2015hw} showed how a robot can learn and utilize
task-specific priors on object properties from observed task-specific
grasps by a human.

To conclude, feature work in computer vision and robotics needs to have
broader scopes in how they define vision. Robotics research using
vision-based approaches for action needs to consider if methods
developed for pure vision are a good match for the action they want the
robot to achieve as primates show a preference for division and
specialization of labor. Robotic grasping research and perhaps robotics
in general needs to take a holistic approach to learning actions.
Actions do not exist in solitude but are part of a complex behavioral
machinery that contains many interdependent parts that provide useful
information about each other. If roboticists can learn to incorporate
these types of holistic perspectives much will be won.

\bibliography{references}{}

\begin{thebibliography}{100}

\bibitem{DBLP:journals/corr/AgrawalNAML16}
Pulkit Agrawal, Ashvin Nair, Pieter Abbeel, Jitendra Malik, and Sergey Levine.
\newblock {Learning to Poke by Poking: Experiential Learning of Intuitive
  Physics}.
\newblock In {\em Neural Information Processing Systems}, pages 5074--5082,
  2016.

\bibitem{Aleotti:2011hc}
Jacopo Aleotti and Stefano Caselli.
\newblock {Part-based robot grasp planning from human demonstration}.
\newblock In {\em IEEE International Conference on Robotics and Automation},
  pages 4554--4560, 2011.

\bibitem{Amedi:2001dc}
Amir Amedi, Rafael Malach, Talma Hendler, Sharon Peled, and Ehud Zohary.
\newblock {Visuo-haptic object-related activation in the ventral visual
  pathway}.
\newblock {\em Nature Neuroscience}, 4(3):324--330, 2001.

\bibitem{Ansuini:2007fn}
Caterina Ansuini, Livia Giosa, Luca Turella, Gianmarco Alto{\`e}, and Umberto
  Castiello.
\newblock {An object for an action, the same object for other actions: effects
  on hand shaping}.
\newblock {\em Experimental Brain Research}, 185(1):111--119, 2007.

\bibitem{Ansuini:2006eu}
Caterina Ansuini, Marco Santello, Stefano Massaccesi, and Umberto Castiello.
\newblock {Effects of End-Goal on Hand Shaping}.
\newblock {\em Journal of Neurophysiology}, 95(4):2456--2465, 2006.

\bibitem{DBLP:conf/corl/AntonovaKSK18}
Rika Antonova, Mia Kokic, Johannes~A Stork, and Danica Kragic.
\newblock {Global Search with Bernoulli Alternation Kernel for Task-oriented
  Grasping Informed by Simulation}.
\newblock In {\em 2nd Annual Conference on Robot Learning}, pages 641--650,
  2018.

\bibitem{Baldassi:2013gu}
Carlo Baldassi, Alireza Alemi-Neissi, Marino Pagan, James~J DiCarlo, Riccardo
  Zecchina, and Davide Zoccolan.
\newblock {Shape Similarity, Better than Semantic Membership, Accounts for the
  Structure of Visual Object Representations in a Population of Monkey
  Inferotemporal Neurons}.
\newblock {\em PLoS Computational Biology}, 9(8):1--20, 2013.

\bibitem{WBarsalou:1999ex}
Lawrence~W Barsalou.
\newblock {Perceptual Symbol Systems}.
\newblock {\em The Behavioral and Brain Sciences}, 22:577--609-- discussion
  610, 1999.

\bibitem{Barsalou:2008ff}
Lawrence~W Barsalou.
\newblock {Grounded Cognition}.
\newblock {\em Annual review of psychology}, 59(1):617--645, 2008.

\bibitem{Barsalou:2016is}
Lawrence~W Barsalou.
\newblock {On Staying Grounded and Avoiding Quixotic Dead Ends}.
\newblock {\em Psychonomic bulletin {\&} review}, 23(4):1122--1142, 2016.

\bibitem{Baugh1262}
Lee~A Baugh, Michelle Kao, Roland~S Johansson, and J~Randall Flanagan.
\newblock {Material evidence: interaction of well-learned priors and
  sensorimotor memory when lifting objects}.
\newblock {\em Journal of Neurophysiology}, 108(5):1262--1269, 2012.

\bibitem{DBLP:journals/trob/BekirogluLJKK11}
Yasemin Bekiroglu, Janne Laaksonen, Jimmy~A J{\o}rgensen, Ville Kyrki, and
  Danica Kragic.
\newblock {Assessing Grasp Stability Based on Learning and Haptic Data}.
\newblock {\em IEEE Transactions on Robotics}, 27(3):616--629, 2011.

\bibitem{Belardinelli:2016ib}
Anna Belardinelli, Madeleine~Y Stepper, and Martin~V Butz.
\newblock {It's in the eyes: Planning precise manual actions before execution}.
\newblock {\em Journal of vision}, 16(1):18--18, 2016.

\bibitem{Bicchi:2000wr}
A~Bicchi and V~Kumar.
\newblock {Robotic Grasping and Contact: A Review}.
\newblock In {\em IEEE International Conference on Robotics and Automation},
  pages 348--353, 2000.

\bibitem{Biederman:1987bx}
Irving Biederman.
\newblock {Recognition-by-components: A theory of human image understanding}.
\newblock {\em Psychological review}, 94(2):115--117, 1987.

\bibitem{Binder:2011wu}
J~R Binder and R~H Desai.
\newblock {The neurobiology of semantic memory}.
\newblock {\em Trends in Cognitive Sciences}, 15(11):527--536, 2011.

\bibitem{DBLP:journals/trob/BohgHSBKSS17}
Jeannette Bohg, Karol Hausman, Bharath Sankaran, Oliver Brock, Danica Kragic,
  Stefan Schaal, and Gaurav~S Sukhatme.
\newblock {Interactive Perception: Leveraging Action in Perception and
  Perception in Action}.
\newblock {\em IEEE Transactions on Robotics}, 33(6):1273--1291, 2017.

\bibitem{6672028}
Jeannette Bohg, A~Morales, Tamim Asfour, and Danica Kragic.
\newblock {Data-Driven Grasp Synthesis - A Survey}.
\newblock {\em IEEE transactions on robotics}, 30(2):289--309, 2014.

\bibitem{Booth:1998tt}
M~C Booth and Edmund~T Rolls.
\newblock {View-invariant representations of familiar objects by neurons in the
  inferior temporal visual cortex.}
\newblock {\em Cerebral Cortex}, 8(6):510--523, 1998.

\bibitem{Buckingham:2009dd}
Gavin Buckingham, Jonathan~S Cant, and Melvyn~A Goodale.
\newblock {Living in A Material World: How Visual Cues to Material Properties
  Affect the Way That We Lift Objects and Perceive Their Weight}.
\newblock {\em Journal of Neurophysiology}, 102(6):3111--3118, 2009.

\bibitem{Bugatus:2017wl}
Lior Bugatus, Kevin~S Weiner, and Kalanit Grill-Spector.
\newblock {Task alters category representations in prefrontal but not
  high-level visual cortex}.
\newblock {\em NeuroImage}, 155:437--449, 2017.

\bibitem{Castiello:2005gq}
Umberto Castiello.
\newblock {The neuroscience of grasping}.
\newblock {\em Nature Reviews Neuroscience}, 6(9):726--736, 2005.

\bibitem{Castiello:2007iy}
Umberto Castiello and C~Begliomini.
\newblock {The Cortical Control of Visually Guided Grasping}.
\newblock {\em The Neuroscientist}, 14(2):157--170, 2007.

\bibitem{Chao:2000bf}
Linda~L Chao and Alex Martin.
\newblock {Representation of manipulable man-made objects in the dorsal
  stream.}
\newblock {\em NeuroImage}, 12(4):478--484, 2000.

\bibitem{Chinellato:2015uj}
Eris Chinellato and Angel~P del Pobil.
\newblock {\em {The Visual Neuroscience of Robotic Grasping}}, volume~28 of
  {\em Achieving Sensorimotor Skills through Dorsal-Ventral Stream
  Integration}.
\newblock Springer, 2016.

\bibitem{Cox:2014km}
David~Daniel Cox.
\newblock {Do we understand high-level vision?}
\newblock {\em Current Opinion in Neurobiology}, 25:187--193, 2014.

\bibitem{Craje:2011eg}
C{\'e}line Craj{\'e}, Jamie~R Lukos, Caterina Ansuini, Andrew~M Gordon, and
  Marco Santello.
\newblock {The effects of task and content on digit placement on a bottle}.
\newblock {\em Experimental Brain Research}, 212(1):119--124, 2011.

\bibitem{Creem:2001ji}
Sarah~H Creem and Dennis~R Proffitt.
\newblock {Grasping objects by their handles: A necessary interaction between
  cognition and action.}
\newblock {\em Journal of Experimental Psychology: Human Perception and
  Performance}, 27(1):218--228, 2001.

\bibitem{Csurka04visualcategorization}
Gabriella Csurka, Christopher~R Dance, Lixin Fan, Jutta Willamowski, and
  C{\'e}dric Bray.
\newblock {Visual categorization with bags of keypoints}.
\newblock {\em European Conference on Computer Vision}, pages 1--22, 2004.

\bibitem{Culham:2004uy}
J~C Culham.
\newblock {Human brain imaging reveals a parietal area specialized for
  grasping}.
\newblock In Nancy Kanwisher and John Duncan, editors, {\em Attention and
  performance XX. Functional brain imaging of visual cognition}, pages
  417--438. Oxford University Press, 2004.

\bibitem{detry2017b}
Renaud Detry, Jeremie Papon, and Larry Matthies.
\newblock {Semantic and Geometric Scene Understanding for Task-oriented
  Grasping of Novel Objects from a Single View}.
\newblock In {\em ICRA Workshop on Learning and control for autonomous
  manipulation systems: the role of dimensionality reduction}, 2017.

\bibitem{DiCarlo:2007hs}
James~J DiCarlo and David~Daniel Cox.
\newblock {Untangling invariant object recognition}.
\newblock {\em Trends in Cognitive Sciences}, 11(8):333--341, 2007.

\bibitem{DiCarlo:2012em}
James~J DiCarlo, Davide Zoccolan, and Nicole~C Rust.
\newblock {How Does the Brain Solve Visual Object Recognition?}
\newblock {\em Neuron}, 73(3):415--434, 2012.

\bibitem{Edelman:1995vd}
Shimon Edelman.
\newblock {Representation, similarity, and the chorus of prototypes.}
\newblock {\em Minds and Machines}, 5(1):45--68, 1995.

\bibitem{Ehrsson:2000vo}
H~Henrik Ehrsson, Anders Fagergren, Tomas Jonsson, G{\"o}ran Westling, Roland~S
  Johansson, and Hans Forssberg.
\newblock {Cortical Activity in Precision- Versus Power-Grip Tasks: An fMRI
  Study}.
\newblock {\em Journal of Neurophysiology}, 83(1):528--536, 2000.

\bibitem{Ekvall:2005ue}
Staffan Ekvall and Danica Kragic.
\newblock {Grasp Recognition for Programming by Demonstration}.
\newblock In {\em IEEE International Conference on Robotics and Automation},
  pages 748--753, 2005.

\bibitem{Ellis:2000uv}
Rob Ellis and Mike Tucker.
\newblock {Micro-affordance: The potentiation of components of action by seen
  objects}.
\newblock {\em British journal of psychology}, 91(4):451--471, 2000.

\bibitem{Fabbri:2016bd}
Sara Fabbri, Kevin~M Stubbs, Rhodri Cusack, and Jody Culham.
\newblock {Disentangling Representations of Object and Grasp Properties in the
  Human Brain}.
\newblock {\em The Journal of Neuroscience}, 36(29):7648--7662, 2016.

\bibitem{Farivar:2009tt}
Reza Farivar.
\newblock {Dorsal--ventral integration in object recognition}.
\newblock {\em Brain Research Reviews}, 61(2):144--153, 2009.

\bibitem{Feix:2014cy}
Thomas Feix, Ian~M Bullock, and Aaron~M Dollar.
\newblock {Analysis of human grasping behavior: correlating tasks, objects and
  grasps.}
\newblock {\em IEEE transactions on haptics}, 7(4):430--441, 2014.

\bibitem{Feix:2014bw}
Thomas Feix, Ian~M Bullock, and Aaron~M Dollar.
\newblock {Analysis of Human Grasping Behavior: Object Characteristics and
  Grasp Type}.
\newblock {\em IEEE transactions on haptics}, 7(3):311--323, 2014.

\bibitem{Felzenszwalb:2010ez}
Pedro~F Felzenszwalb, Ross Girshick, David~A McAllester, and Deva Ramanan.
\newblock {Object Detection with Discriminatively Trained Part-Based Models.}
\newblock {\em IEEE Transactions on Pattern Analysis and Machine Intelligence},
  32(9):1627--1645, 2010.

\bibitem{Flanagan:2009ux}
J~Randall Flanagan and Roland~S Johansson.
\newblock {Sensory control of object manipulation}.
\newblock In {\em Sensorimotor Control of Grasping: Physiology and
  Pathophysiology}, pages 141--160. Cambridge University Press, 2009.

\bibitem{Gallese:2005fl}
Vittorio Gallese and George Lakoff.
\newblock {The Brain's concepts: the role of the Sensory-motor system in
  conceptual knowledge.}
\newblock {\em Cognitive neuropsychology}, 22(3):455--479, 2005.

\bibitem{Gallivan:2017jz}
Jason~P Gallivan, Jonathan~S Cant, Melvyn~A Goodale, and J~Randall Flanagan.
\newblock {Representation of Object Weight in Human Ventral Visual Cortex}.
\newblock {\em Current Biology}, 24(16):1866--1873, 2017.

\bibitem{Gao:2016vs}
Yang Gao, Lisa~Anne Hendricks, Katherine~J Kuchenbecker, and Trevor~J Darrell.
\newblock {Deep Learning for Tactile Understanding From Visual and Haptic
  Data}.
\newblock In {\em IEEE International Conference on Robotics and Automation},
  pages 536--543, 2016.

\bibitem{Goodale:2011ff}
Melvyn~A Goodale.
\newblock {Transforming vision into action}.
\newblock {\em Vision research}, 51(13):1567--1587, 2011.

\bibitem{Goodale:1992gq}
Melvyn~A Goodale and A~David Milner.
\newblock {Separate visual pathways for perception and action}.
\newblock {\em Trends in Neurosciences}, 15(1):20--25, 1992.

\bibitem{Grafton:2010jo}
Scott~T Grafton.
\newblock {The cognitive neuroscience of prehension: recent developments}.
\newblock {\em Experimental Brain Research}, 204(4):475--491, 2010.

\bibitem{Grefkes:2005gf}
Christian Grefkes and Gereon~R Fink.
\newblock {The functional organization of the intraparietal sulcus in humans
  and monkeys}.
\newblock {\em Journal of Anatomy}, 207(1):3--17, 2005.

\bibitem{Grezes:2003bz}
J~Gr{\`e}zes, Mike Tucker, J~Armony, Rob Ellis, and R~E Passingham.
\newblock {Objects automatically potentiate action: an fMRI study of implicit
  processing}.
\newblock {\em The European journal of neuroscience}, 17(12):2735--2740, 2003.

\bibitem{Harel:2014cp}
A~Harel, D~J Kravitz, and C~I Baker.
\newblock {Task context impacts visual object processing differentially across
  the cortex}.
\newblock {\em Proceedings of the National Academy of Sciences of the United
  States of America}, 111(10):962--971, 2014.

\bibitem{Harnad:1999uv}
Stevan Harnad.
\newblock {The Symbol Grounding Problem}.
\newblock {\em Physica D. Nonlinear phenomena}, 42:335--346, 1990.

\bibitem{Herbort:2014dj}
Oliver Herbort, Martin~V Butz, and Wilfried Kunde.
\newblock {The contribution of cognitive, kinematic, and dynamic factors to
  anticipatory grasp selection}.
\newblock {\em Experimental Brain Research}, 232(6):1677--1688, 2014.

\bibitem{Hermens:2014jl}
Frouke Hermens, Daniel Kral, and David~A Rosenbaum.
\newblock {Limits of end-state planning}.
\newblock {\em Acta psychologica}, 148:148--162, 2014.

\bibitem{Hjelm:2014uc}
Martin Hjelm, Renaud Detry, Carl~Henrik Ek, and Danica Kragic.
\newblock {Representations for cross-task, cross-object grasp transfer.}
\newblock {\em IEEE International Conference on Robotics and Automation}, pages
  5699--5704, 2014.

\bibitem{Hjelm:2015hw}
Martin Hjelm, Carl~Henrik Ek, Renaud Detry, and Danica Kragic.
\newblock {Learning Human Priors for Task-Constrained Grasping.}
\newblock In {\em International Conference on Computer Vision Systems}, pages
  207--217, 2015.

\bibitem{Isik:2016ts}
Leyla Isik, Andrea Tacchetti, and Tomaso Poggio.
\newblock {A fast, invariant representation for human action in the visual
  system}.
\newblock {\em Journal of Neurophysiology}, 119(2):631--640, 2018.

\bibitem{James2463}
Thomas~W James, Jody Culham, G~Keith Humphrey, A~David Milner, and Melvyn~A
  Goodale.
\newblock {Ventral occipital lesions impair object recognition but not
  object-directed grasping: an fMRI study}.
\newblock {\em Brain}, 126(11):2463--2475, 2003.

\bibitem{James:2002gf}
Thomas~W James, G~Keith Humphrey, Joseph~S Gati, Ravi~S Menon, and Melvyn~A
  Goodale.
\newblock {Differential Effects of Viewpoint on Object-Driven Activation in
  Dorsal and Ventral Streams}.
\newblock {\em Neuron}, 35(4):793--801, 2002.

\bibitem{Jeannerod:1999ke}
M~Jeannerod.
\newblock {Visuomotor channels: Their integration in goal-directed prehension}.
\newblock {\em Human Movement Science}, 18(2-3):201--218, 1999.

\bibitem{Jenmalm:2000ui}
Per Jenmalm, Seth Dahlstedt, and Roland~S Johansson.
\newblock {Visual and Tactile Information About Object-Curvature Control
  Fingertip Forces and Grasp Kinematics in Human Dexterous Manipulation}.
\newblock {\em Journal of Neurophysiology}, 84(6):2984--2997, 2000.

\bibitem{Johansson6917}
Roland~S Johansson, G{\"o}ran Westling, Anders B{\"a}ckstr{\"o}m, and J~Randall
  Flanagan.
\newblock {Eye-Hand Coordination in Object Manipulation}.
\newblock {\em The Journal of Neuroscience}, 21(17):6917--6932, 2001.

\bibitem{7759657}
E~Johns, S~Leutenegger, and A~J Davison.
\newblock {Deep learning a grasp function for grasping under gripper pose
  uncertainty}.
\newblock In {\em IEEE/RSJ International Conference on Intelligent Robots and
  Systems}, pages 4461--4468, 2016.

\bibitem{JohnsonFrey:2004ek}
Scott Johnson~Frey, Michael McCarty, and Rachel Keen.
\newblock {Reaching beyond spatial perception: Effects of intended future
  actions on visually guided prehension}.
\newblock {\em Visual Cognition}, 11(2-3):371--399, 2004.

\bibitem{Kiefer:2012bi}
Markus Kiefer and Friedemann Pulverm{\"u}ller.
\newblock {Conceptual representations in mind and brain: Theoretical
  developments, current evidence and future directions}.
\newblock {\em Cortex}, 48(7):805--825, 2012.

\bibitem{DBLP:conf/humanoids/KokicSHK17}
Mia Kokic, Johannes~A Stork, Joshua~A Haustein, and Danica Kragic.
\newblock {Affordance detection for task-specific grasping using deep
  learning}.
\newblock In {\em IEEE-RAS International Conference on Humanoid Robotics},
  pages 91--98, 2017.

\bibitem{Kravitz:2013uk}
D~J Kravitz, K~S Saleem, and C~I Baker.
\newblock {The ventral visual pathway: an expanded neural framework for the
  processing of object quality}.
\newblock {\em Trends in Cognitive Sciences}, 17(1):26--49, 2013.

\bibitem{Kravitz:2011vy}
Dwight~J Kravitz, Kadharbatcha~S Saleem, Chris~I Baker, and Mortimer Mishkin.
\newblock {A new neural framework for visuospatial processing}.
\newblock {\em Nature Reviews Neuroscience}, 12:217--230, 2011.

\bibitem{Krug:2011fg}
K~Krug and Andrew~J Parker.
\newblock {Neurons in Dorsal Visual Area V5/MT Signal Relative Disparity}.
\newblock {\em The Journal of Neuroscience}, 31(49):17892--17904, 2011.

\bibitem{Kruger:2013wg}
Norbert Kr{\"u}ger, Peter Janssen, Sinan Kalkan, Markus Lappe, Ales Leonardis,
  Justus~H Piater, Antonio~Jose Rodriguez-Sanchez, and Laurenz Wiskott.
\newblock {Deep Hierarchies in the Primate Visual Cortex - What Can We Learn
  for Computer Vision?}
\newblock {\em IEEE Transactions on Pattern Analysis and Machine Intelligence},
  35(8):1847--1871, 2013.

\bibitem{Lebedev:2002kh}
M~A Lebedev and S~P Wise.
\newblock {Insights into Seeing and Grasping: Distinguishing the Neural
  Correlates of Perception and Action}.
\newblock {\em Behavioral and Cognitive Neuroscience Reviews}, 1(2):108--129,
  2002.

\bibitem{Lenz:2013uz}
Ian Lenz, Honglak Lee, and Ashutosh Saxena.
\newblock {Deep Learning for Detecting Robotic Grasps}.
\newblock {\em The International Journal of Robotics Research}, 34:705--724,
  2015.

\bibitem{DBLP:journals/corr/LevinePKQ16}
Sergey Levine, Peter Pastor, Alex Krizhevsky, and Deirdre Quillen.
\newblock {Learning Hand-Eye Coordination for Robotic Grasping with Deep
  Learning and Large-Scale Data Collection}.
\newblock {\em The International Journal of Robotics Research}, 37:421--436,
  2018.

\bibitem{6943027}
M~Li, Y~Bekiroglu, Danica Kragic, and Aude~G Billard.
\newblock {Learning of grasp adaptation through experience and tactile
  sensing}.
\newblock In {\em IEEE/RSJ International Conference on Intelligent Robots and
  Systems}, pages 3339--3346, 2014.

\bibitem{Logothetis:1999ub}
N~K Logothetis.
\newblock {Vision: a window on consciousness.}
\newblock {\em Scientific American}, 281(5):69--75, 1999.

\bibitem{Logothetis:1996tr}
N~K Logothetis and D~L Sheinberg.
\newblock {Visual Object Recognition}.
\newblock {\em Annual Review of Neuroscience}, 19(1):577--621, 1996.

\bibitem{MacKenzie:1994wh}
C~L MacKenzie and T~Iberall.
\newblock {\em {The Grasping Hand}}.
\newblock Advances in Psychology. Elsevier, 1994.

\bibitem{Mahon:2008fv}
Bradford~Z Mahon and Alfonso Caramazza.
\newblock {A critical look at the embodied cognition hypothesis and a new
  proposal for grounding conceptual content}.
\newblock {\em Journal of Physiology-Paris}, 102(1-3):59--70, 2008.

\bibitem{Martin:2007ck}
Alex Martin.
\newblock {The Representation of Object Concepts in the Brain}.
\newblock {\em Annual review of psychology}, 58(1):25--45, 2007.

\bibitem{Martin:2001cp}
Alex Martin and Linda~L Chao.
\newblock {Semantic memory and the brain: structure and processes}.
\newblock {\em Current Opinion in Neurobiology}, 11(2):194--201, 2001.

\bibitem{McIntosh:2008jd}
Robert~D McIntosh and Gavin Lashley.
\newblock {Matching boxes: Familiar size influences action programming}.
\newblock {\em Neuropsychologia}, 46(9):2441--2444, 2008.

\bibitem{Miller:2003ev}
A~T Miller, S~Knoop, H~I Christensen, and Peter~K Allen.
\newblock {Automatic grasp planning using shape primitives}.
\newblock In {\em IEEE International Conference on Robotics and Automation},
  pages 1824--1829, 2003.

\bibitem{Milner:2003io}
A~David Milner, H~C Dijkerman, Robert~D McIntosh, Y~Rossetti, and L~Pisella.
\newblock {Delayed reaching and grasping in patients with optic ataxia}.
\newblock In {\em Neural Control of Space Coding and Action Production}, pages
  225--242. Elsevier, 2003.

\bibitem{Napier:1956tn}
J~R Napier.
\newblock {The prehensile movements of the human hand.}
\newblock {\em The Journal of Bone and Joint Surgery. British volume},
  38-B(4):902--913, 1956.

\bibitem{Neri:2004ix}
P~Neri.
\newblock {A Stereoscopic Look at Visual Cortex}.
\newblock {\em Journal of Neurophysiology}, 93(4):1823--1826, 2004.

\bibitem{Nikandrova:2015uu}
E~Nikandrova and Ville Kyrki.
\newblock {Category-based task specific grasping}.
\newblock {\em Robotics and Autonomous Systems}, 70:25--35, 2015.

\bibitem{Palmeri:2004bi}
Thomas~J Palmeri and Isabel Gauthier.
\newblock {Visual object understanding}.
\newblock {\em Nature Reviews Neuroscience}, 5(4):291--303, 2004.

\bibitem{Peissig:2007dt}
Jessie~J Peissig and Michael~J Tarr.
\newblock {Visual object recognition: do we know more now than we did 20 years
  ago?}
\newblock {\em Annual review of psychology}, 58:75--96, 2007.

\bibitem{DBLP:journals/corr/PintoGHPG16}
Lerrel Pinto, Dhiraj Gandhi, Yuanfeng Han, Yong-Lae Park, and Abhinav Gupta.
\newblock {The Curious Robot: Learning Visual Representations via Physical
  Interactions}.
\newblock In {\em European conference on Computer Vision}, pages 3--18, 2016.

\bibitem{Pinto:2008gj}
Nicolas Pinto, David~Daniel Cox, and James~J DiCarlo.
\newblock {Why is Real-World Visual Object Recognition Hard?}
\newblock {\em PLoS Computational Biology}, 4(1):1--6, 2008.

\bibitem{Pulvermuller:2014jr}
Friedemann Pulverm{\"u}ller, Rachel~L Moseley, Natalia Egorova, Zubaida
  Shebani, and V{\'e}ronique Boulenger.
\newblock {Motor cognition--motor semantics: Action perception theory of
  cognition and communication}.
\newblock {\em Neuropsychologia}, 55:71--84, 2014.

\bibitem{10.1371/journal.pone.0012608}
Jenny C~A Read, Graeme~P Phillipson, Ignacio Serrano-Pedraza, A~David Milner,
  and Andrew~J Parker.
\newblock {Stereoscopic Vision in the Absence of the Lateral Occipital Cortex}.
\newblock {\em PLoS One}, 5(9):1--14, 2010.

\bibitem{Redmon:2015wj}
Joseph Redmon and Anelia Angelova.
\newblock {Real-time grasp detection using convolutional neural networks.}
\newblock In {\em IEEE International Conference on Robotics and Automation},
  pages 1316--1322, 2015.

\bibitem{Riesenhuber:2000bw}
Maximilian Riesenhuber and Tomaso Poggio.
\newblock {Models of object recognition}.
\newblock {\em Nature Neuroscience}, 3:1199--1204, 2000.

\bibitem{Rizzolatti1988}
Giacomo Rizzolatti, R~Camarda, Leonardo Fogassi, M~Gentilucci, Giuseppe
  Luppino, and M~Matelli.
\newblock {Functional organization of inferior area 6 in the macaque monkey}.
\newblock {\em Experimental Brain Research}, 71(3):491--507, 1988.

\bibitem{Rizzolatti:1996fo}
Giacomo Rizzolatti, Luciano Fadiga, Vittorio Gallese, and Leonardo Fogassi.
\newblock {Premotor cortex and the recognition of motor actions}.
\newblock {\em Brain research. Cognitive brain research}, 3(2):131--141, 1996.

\bibitem{Rochat:2010hy}
Magali~J Rochat, Fausto Caruana, Ahmad Jezzini, Ludovic Escola, Irakli
  Intskirveli, Franck Grammont, Vittorio Gallese, Giacomo Rizzolatti, and
  Maria~Alessandra Umilt{\`a}.
\newblock {Responses of mirror neurons in area F5 to hand and tool grasping
  observation}.
\newblock {\em Experimental Brain Research}, 204(4):605--616, 2010.

\bibitem{DBLP:conf/iccv/RogezSR15}
Gr~e~gory Rogez, James~Steven Supancic, III, and Deva Ramanan.
\newblock {Understanding Everyday Hands in Action from RGB-D Images}.
\newblock In {\em IEEE International Conference on Computer Vision}, pages
  3889--3897, 2015.

\bibitem{Rolls:2012dt}
Edmund~T Rolls.
\newblock {Invariant Visual Object and Face Recognition - Neural and
  Computational Bases, and a Model, VisNet.}
\newblock {\em Frontiers in Computational Neuroscience}, 6:35, 2012.

\bibitem{Rolls:2007gu}
Edmund~T Rolls and Simon~M Stringer.
\newblock {Invariant Global Motion Recognition in the Dorsal Visual System: A
  Unifying Theory}.
\newblock {\em Neural Computation}, 19(1):139--169, 2007.

\bibitem{Rosenbaum:2012ki}
David~A Rosenbaum, Kate~M Chapman, Matthias Weigelt, Daniel~J Weiss, and
  Robrecht van~der Wel.
\newblock {Cognition, action, and object manipulation.}
\newblock {\em Psychological Bulletin}, 138(5):924--946, 2012.

\bibitem{10.1371/journal.pone.0025203}
Luisa Sartori, Elisa Straulino, and Umberto Castiello.
\newblock {How Objects Are Grasped: The Interplay between Affordances and
  End-Goals}.
\newblock {\em PLoS One}, 6(9):1--10, 2011.

\bibitem{Saxena:2008wn}
Ashutosh Saxena, Justin Driemeyer, and Andrew~Y. Ng.
\newblock {Robotic Grasping of Novel Objects using Vision}.
\newblock {\em The International Journal of Robotics Research}, 27(2):157--173,
  2008.

\bibitem{Schenk:2010gm}
Thomas Schenk.
\newblock {Visuomotor robustness is based on integration not segregation}.
\newblock {\em Vision research}, 50(24):2627--2632, 2010.

\bibitem{Schenk:2011uh}
Thomas Schenk, V~Franz, and N~Bruno.
\newblock {Vision-for-perception and vision-for-action: Which model is
  compatible with the available psychophysical and neuropsychological data?}
\newblock {\em Vision research}, 51(8):812--818, 2011.

\bibitem{Schenk:2010jq}
Thomas Schenk and Robert~D McIntosh.
\newblock {Do we have independent visual streams for perception and action?}
\newblock {\em Cognitive Neuroscience}, 1(1):52--62, 2010.

\bibitem{Serre:2007ir}
T~Serre, Aude Oliva, and Tomaso Poggio.
\newblock {A feedforward architecture accounts for rapid categorization}.
\newblock {\em Proceedings of the National Academy of Sciences of the United
  States of America}, 104(15):6424--6429, 2007.

\bibitem{Serre:2007kq}
T~Serre, L~Wolf, Stanley Bileschi, Maximilian Riesenhuber, and Tomaso Poggio.
\newblock {Robust Object Recognition with Cortex-Like Mechanisms}.
\newblock {\em IEEE Transactions on Pattern Analysis and Machine Intelligence},
  29(3):411--426, 2007.

\bibitem{Song:2010gh}
Dan Song, Kai Huebner, Ville Kyrki, and Danica Kragic.
\newblock {Learning task constraints for robot grasping using graphical
  models.}
\newblock In {\em IEEE/RSJ International Conference on Intelligent Robots and
  Systems}, pages 1579--1585, 2010.

\bibitem{Tarr:2002bn}
Michael~J Tarr and Quoc~C Vuong.
\newblock {\em {Visual Object Recognition}}.
\newblock Stevens' Handbook of Experimental Psychology. John Wiley {\&} Sons,
  Inc., 2002.

\bibitem{Tucker:2001ey}
Mike Tucker and Rob Ellis.
\newblock {The potentiation of grasp types during visual object
  categorization}.
\newblock {\em Visual Cognition}, 8(6):769--800, 2001.

\bibitem{Ungerleider:1982uz}
Leslie Ungerleider and Mortimer Mishkin.
\newblock {Two cortical visual systems}.
\newblock In {\em Analysis of Visual Behavior}, pages 549--586. MIT Press,
  1982.

\bibitem{VaziriPashkam:2017cu}
Maryam Vaziri-Pashkam and Yaoda Xu.
\newblock {Goal-directed visual processing differentially impacts human ventral
  and dorsal visual representations}.
\newblock {\em The Journal of Neuroscience}, pages 3392--16, 2017.

\bibitem{Wallis:1997er}
Guy Wallis and Edmund~T Rolls.
\newblock {Invariant Face And Object Recognition In The Visual System}.
\newblock {\em Progress in Neurobiology}, 51(2):167--194, 1997.

\bibitem{Zachariou:2014cl}
Valentinos Zachariou, Roberta Klatzky, and Marlene Behrmann.
\newblock {Ventral and Dorsal Visual Stream Contributions to the Perception of
  Object Shape and Object Location}.
\newblock {\em Journal of Cognitive Neuroscience}, 26(1):189--209, 2014.

\end{thebibliography}
\bibliographystyle{plain}

\end{document}